\documentclass{article}
\usepackage{iclr2027_conference,times}
\usepackage[hyphens]{url}
\usepackage{graphicx}
\usepackage{booktabs}
\usepackage{amsmath}
\usepackage{amssymb}
\usepackage{multirow}
\usepackage{xcolor}
\usepackage{colortbl}
\usepackage{placeins}
\usepackage{hyperref}
\newcommand{\srcverified}{\textsuperscript{\ensuremath{\dagger}}}

\newcommand{\datasetname}{\textsc{ConClick-Data}}
\newcommand{\modelname}{\textsc{ConClick}}
\newcommand{\method}{\textsc{ReClick}}

\graphicspath{{./}{figures/}{Figures/}}

\title{ConClick: Element-wise Contrastive \\ Learning Improves GUI Grounding}
\author{Ruotian Zhang, Yuhui Zhang, Junzhe Huang, Yu Xin, Hehe Fan, Xiaoxiao Sun\\
\{ruotianzh, xxsunzrt\}@gmail.com
}

\iclrfinalcopy 
\begin{document} 
\maketitle

\begin{abstract}
    GUI grounding, the task of locating the interface element specified by a user instruction, is a fundamental capability of GUI agents. However, existing grounding models remain error-prone, often causing downstream task failures. We systematically evaluate 10 grounding models across five datasets and identify a major error pattern: most incorrect clicks hit non-target interface elements rather than the background, with different models repeatedly selecting the same distractors. Motivated by these findings, we propose \modelname{}, a contrastive learning framework that teaches models to distinguish the target element from plausible distractors. We construct \datasetname{}, a dataset containing 4,292 challenging target–distractor pairs, and train the model using complementary listwise ranking and pairwise margin objectives. \modelname{} improves the grounding accuracy of Qwen3-VL-8B from 54.71\% to 66.03\% on ScreenSpot-Pro, from 83.72\% to 86.28\% on MMBench-GUI L2, and from 63.33\% to 70.20\% on OSWorld-G, outperforming prior methods built on the same or similarly sized backbones. We further introduce \method{}, an inference-time strategy that revisits promising regions and compares predictions across competing candidates, yielding an additional 2.96 percentage-point improvement in average accuracy across the three benchmarks. Together, our analysis, dataset, and methods provide both empirical insights and practical tools for building more reliable GUI agents. Data and code are available at \url{https://github.com/zhangruotian/ConClick}.
\end{abstract} 

\section{Introduction}
\begin{figure*}[t]
    \centering
    \includegraphics[width=\textwidth]{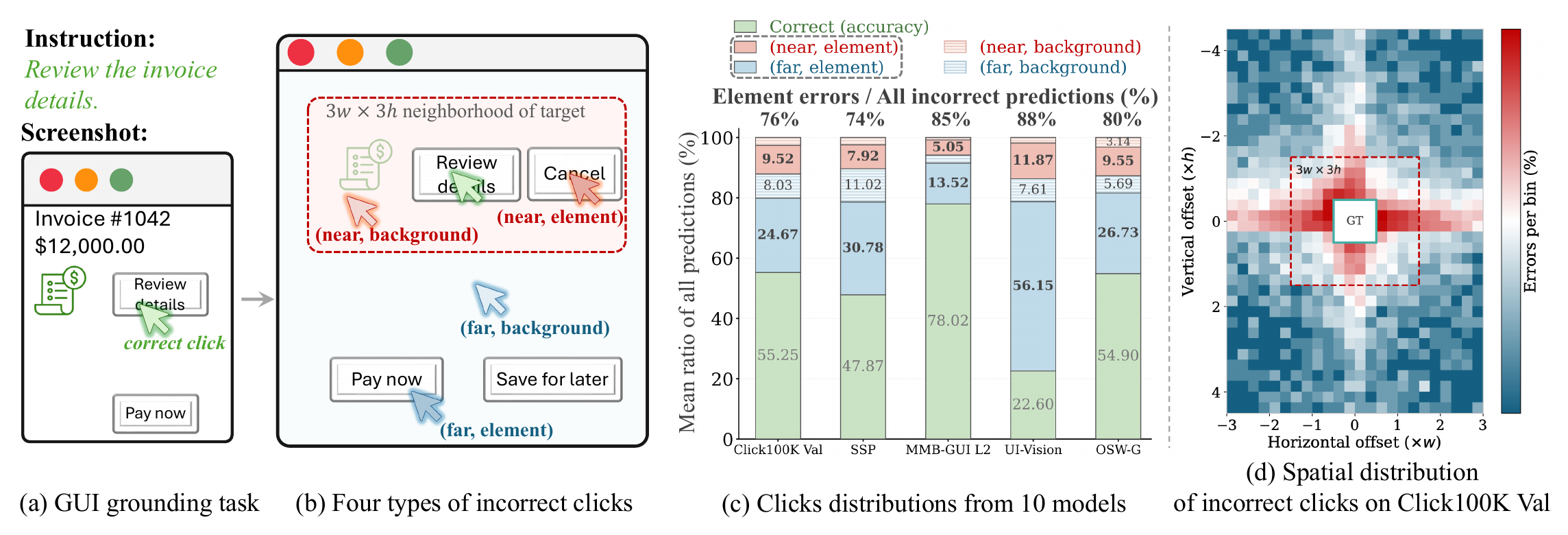}
    \vspace{-5mm}
    \caption{\textbf{Characterizing incorrect clicks.} (a) GUI grounding predicts a click within the target region given an instruction and a screenshot. (b) Incorrect clicks are categorized by whether they fall on an element or background and inside or outside the target-centered $3w\times3h$ neighborhood. (c) Mean accuracy and four error rates across 10 models on five datasets, each using all predictions as the denominator. (d) Spatial distribution of incorrect clicks pooled across 10 models on Click100K Val. Offsets are normalized by target width and height; colors show the percentage of all predictions per $0.25\times0.25$ bin on a logarithmic scale. Red indicates more errors.}
    \label{fig:diagnostic_overview}
\vspace{-3mm}
\end{figure*}

GUI grounding maps instructions and screenshots to click locations\footnote{Throughout this paper, we use ``click'' to refer to the predicted location.}, providing a core capability for GUI agents. Substantial effort has been devoted to improving GUI grounding through stronger multimodal backbones~\citep{qwen3vl,internvl35}, larger datasets~\citep{uground,osatlas,osworldg,uitars,opencua}, and reinforcement learning~\citep{guig1,segui,guig2,infiguig1,uivenus}. Inference methods yield further improvements by searching regions, enlarging crops, repeating predictions, or re-encoding selected regions~\citep{regionfocus,mvp,zoomclick,reprefill,guirc}. All these methods are evaluated using click-point accuracy, the percentage of correct clicks, defined as those falling within the annotated target region (Fig.~\ref{fig:diagnostic_overview}(a)). However, this criterion treats all off-target clicks equally and does not reveal where incorrect clicks land. For example, when instructed to review the invoice details in Fig.~\ref{fig:diagnostic_overview}, clicking ``Pay now'' instead of ``Review details'' can initiate an unintended payment, whereas clicking the background may have no effect. Wrong-button clicks may also delete files or expose confidential information, potentially causing irreversible harm that accuracy alone does not capture. Distinguishing error types is therefore important for assessing the potential consequences of grounding failures.

To bridge this gap, we examine incorrect clicks along two dimensions: element selection and spatial proximity. The element-level dimension distinguishes clicks on a non-target GUI element from those on the background. The spatial dimension distinguishes clicks near or far from the target element according to whether they fall inside or outside a target-centered neighborhood.\footnote{The neighborhood is a target-centered rectangle, such as one of size $3w\times3h$, where $w$ and $h$ are the target width and height. Near and far are relative to this neighborhood, so far need not imply a large absolute distance.} Combining these dimensions yields four error categories: \emph{near-element}, \emph{near-background}, \emph{far-element}, and \emph{far-background} (Fig.~\ref{fig:diagnostic_overview}(b)). We use this categorization to characterize incorrect clicks, with a particular focus on \textbf{element selection errors}. We evaluate 10 models across five datasets: a validation set of Click100K~\citep{gelato2025} and four test sets, ScreenSpot-Pro~\citep{screenspotpro}, MMBench-GUI L2~\citep{mmbenchgui}, UI-Vision~\citep{uivision}, and OSWorld-G~\citep{osworldg}.

Fig.~\ref{fig:diagnostic_overview}(c) shows accuracy and the four error rates over all predictions, averaged across 10 models per dataset. These distributions reveal consistent error patterns across datasets. Specifically, \textcolor[HTML]{E5AAA1}{\rule{0.85em}{0.55em}}~\emph{near-element} and \textcolor[HTML]{A8C6D6}{\rule{0.85em}{0.55em}}~\emph{far-element} errors, i.e., clicks on non-target GUI elements, together account for approximately 74\%--88\% of all incorrect predictions across the five datasets. The predominance of element errors also holds when the 10 models are analyzed separately (Fig.~\ref{fig:app_per_model_distributions}). We conduct further analysis of incorrect element selection. Results show that element coverage and uniform random clicking cannot explain the predominance of element errors. We refer to non-target GUI elements selected by incorrect predictions as distractor elements and observe that different models can select the same distractor for the same instruction and screenshot. Our spatial analysis also shows that errors are denser inside the target neighborhood despite being more numerous outside it (Fig.~\ref{fig:diagnostic_overview}(d)).

These results on element selection suggest that GUI models still struggle to distinguish the target from distractor elements, motivating us to explore supervision that explicitly accounts for distractors. Traditional coordinate supervision is effective for learning to predict target coordinates, but does not explicitly contrast the target with the distractor it is likely to select.
Existing training examples with target-only grounding annotations do not provide the corresponding distractor elements needed for this supervision. To support it, we first construct \datasetname{}, a small dataset of about 8k examples from Click100K: half contain target--distractor annotations, and half are ordinary grounding examples sampled across source categories to maintain data diversity. We use two GUI models, GTA1-7B~\citep{gta1} and GUI-Libra-8B~\citep{guilibra}, to identify observed distractors by matching their incorrect predictions to non-target GUI elements.

For training, we introduce \modelname{}, an element-wise contrastive training strategy that learns to prefer the target over observed distractors for the same instruction. Specifically, a listwise ranking loss encourages the model to rank the target above candidate distractors, while a pairwise margin loss encourages a stronger preference for the target over the annotated distractor. We train the model with a coordinate loss on all examples and an additional selection loss combining these two terms on target--distractor examples, directly addressing the observed element selection errors.

As a natural extension, we introduce \method{} to further improve target selection at inference time by comparing competing candidate clicks. It uses the region preferences learned during training to select regions for revisiting and obtain candidate predictions. Motivated in part by the higher error density near the target, it also examines the area around the initial click. Each candidate is evaluated by the support it receives from other views, measured by cross-view likelihood, and its spatial agreement with other candidates. We combine these two scores with equal weights to select the final prediction, using the same comparison rule on all benchmarks.

Experiments show that \modelname{} improves grounding accuracy across the models and datasets. On ScreenSpot-Pro, it raises Qwen3-VL-8B accuracy from 54.71\% to 66.03\%, with \method{} further improving it to 72.36\%. Together, \modelname{} and \method{} reduce the element error rate across all three benchmarks. On ScreenSpot-Pro, \method{} also reduces near-target error density and achieves higher accuracy than MVP or ZoomClick applied to the same trained model.

Our contributions are summarized as follows: 
\setlength{\leftmargini}{20pt}  
\begin{itemize}
\item We dissect GUI grounding errors across models and datasets, revealing that most incorrect clicks hit other elements and that different models repeatedly select the same distractors.
\item We introduce \modelname{}, an element-wise contrastive training strategy that strengthens discrimination between target and distractor elements. We also introduce \method{} to guide region revisits and compare candidate clicks at inference time.
\item Experiments show that element-wise contrastive training improves grounding and reduces clicks on non-target elements across models and datasets, with improvements from \method{}.
\end{itemize}

\section{Dissecting Incorrect Clicks in GUI Grounding}
\label{sec:discovery_analysis}

\subsection{Error Decomposition Setup}
\label{sec:error_decomposition_setup}

\noindent\textbf{Model and data.} We evaluate 10 GUI grounding models: GTA1-7B~\citep{gta1}, GUI-Libra-8B~\citep{guilibra}, Qwen3-VL-8B~\citep{qwen3vl}, V2P-7B~\citep{v2p}, OpenCUA-7B~\citep{opencua}, GUI-G2-7B~\citep{guig2}, UI-TARS-1.5-7B~\citep{uitars15}, SE-GUI-7B~\citep{segui}, GUI-Actor-7B~\citep{guiactor}, and InternVL3.5-8B~\citep{internvl35}, and closely reproduce reported accuracies using the official model configurations. Finally, 161,640 predictions are collected across the 10 models on a 5,000-example validation set of Click100K and four benchmarks: ScreenSpot-Pro, MMBench-GUI L2, UI-Vision, and OSWorld-G. For OSWorld-G, we use only target-bearing examples, excluding refusal examples. In addition to the ground-truth target annotations, we use OmniParser-v2~\citep{omniparserv2} to detect GUI element regions in each screenshot. Manual inspection indicates that the detections are generally reliable (App.~\ref{sec:app_prediction_corpus}). We assign an incorrect click to the element category if it falls within a detected GUI element region and to the background category otherwise.

\subsection{Error Patterns and Further Analysis}   
\label{sec:error_categories_analysis}

\noindent\textbf{Incorrect element selections dominate across individual models.} Across all 50 model--dataset combinations, near-element and far-element errors together account for 63.68\%--95.43\% of incorrect predictions, with far-element consistently the largest error category (see App. Fig.~\ref{fig:app_per_model_distributions} for all results). The predominance of these errors is concerning because \textbf{clicking the wrong element can trigger harmful actions that are difficult or impossible to reverse}. For example, an agent may submit a payment instead of reviewing an invoice, permanently delete an important file, or send confidential information to the wrong recipient. These actions can cause financial loss, data loss, or privacy breaches, and correcting a later prediction may not undo the damage. A background click may leave the interface unchanged, yet click-point accuracy counts both outcomes as equally incorrect and does not capture this difference in potential harm.

\begin{figure*}[t]
    \centering
    \includegraphics[width=\textwidth]{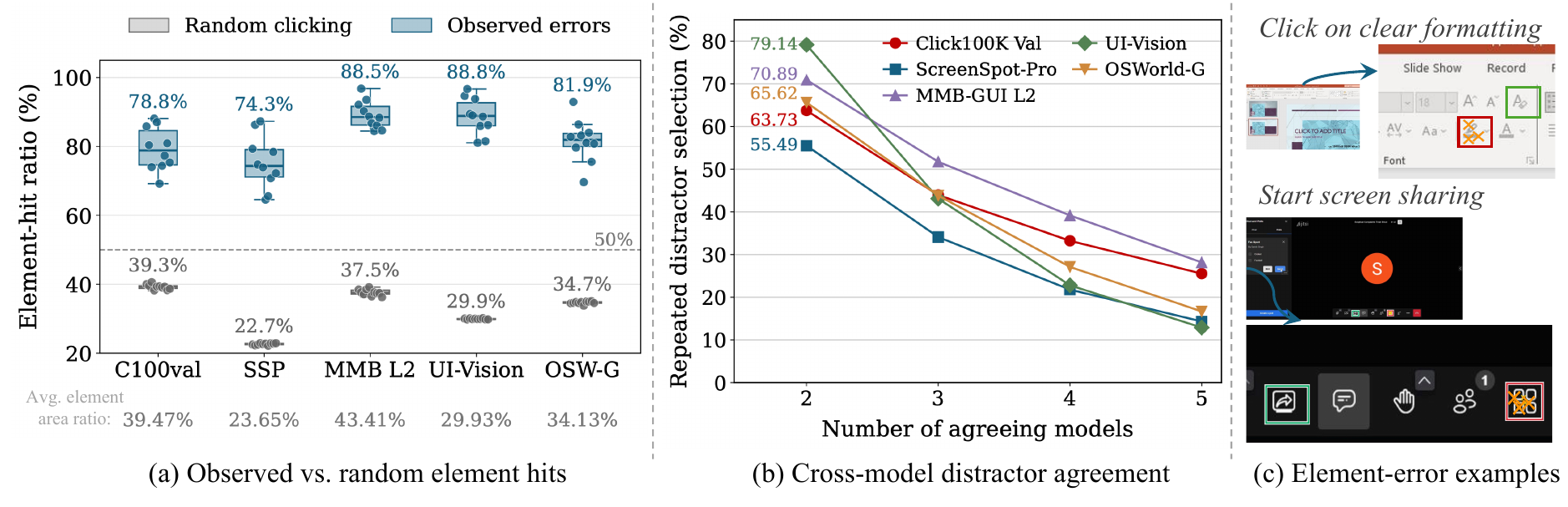}
       \vspace{-5mm}
    \caption{\textbf{Further analysis of incorrect clicks.} (a) Observed versus random element-hit proportions across 10 models and five datasets. Points represent models; random click expectations vary with each model's error set. Box lines and labels show medians. Bottom percentages show dataset-wide mean element coverage. (b) Agreement of at least $k$ models on the same distractor, among examples with at least $k$ incorrect predictions. (c) Element-error examples. Green boxes mark targets, red boxes mark distractors, and orange crosses indicate incorrect element clicks from different models.}
    \label{fig:diagnostic_more}
       \vspace{-3mm}
\end{figure*}

\noindent\textbf{Element coverage and uniform random clicking cannot explain the predominance of element errors.} For each model, we analytically compute the random click expectation as the mean fraction of non-target screen area covered by GUI elements over its incorrect examples, counting overlaps only once. Across the five datasets, median observed element-hit proportions range from 74.3\% to 88.8\%, compared with 22.7\%--39.3\% under uniform random clicking (Fig.~\ref{fig:diagnostic_more}(a)). Across all 50 model--dataset combinations, the observed proportion of element errors exceeds 50\%, while the random-click expectation remains below 50\%.

\noindent\textbf{Different models select the same distractors.}
Fig.~\ref{fig:diagnostic_more}(b) summarizes the results across the five datasets: 55.49\%--79.14\% of examples with at least two incorrect predictions contain a distractor selected by multiple models. The proportions decrease as more models are required to agree, but among examples with at least five incorrect predictions, 12.88\%--28.15\% contain a distractor selected by at least five models. Fig.~\ref{fig:diagnostic_more}(c) illustrates two such cases of different models selecting the same distractor. For the Qwen3-VL-8B error cases from Click100K Val examined in Fig.~\ref{fig:near_error_density}(b), attention scores do not consistently favor the target over the distractor.

These findings motivate explicitly contrasting the target with observed distractors to strengthen target selection. Section~\ref{sec:selectground_method} introduces \modelname{}, an element-wise contrastive training strategy, and \method{} as a natural inference-time extension.

\section{Method}
\label{sec:selectground_method}

\subsection{Element-wise Contrastive Training: \modelname{}}
\label{sec:training_observed}

\begin{figure*}[t]
    \centering
    \includegraphics[width=\textwidth]{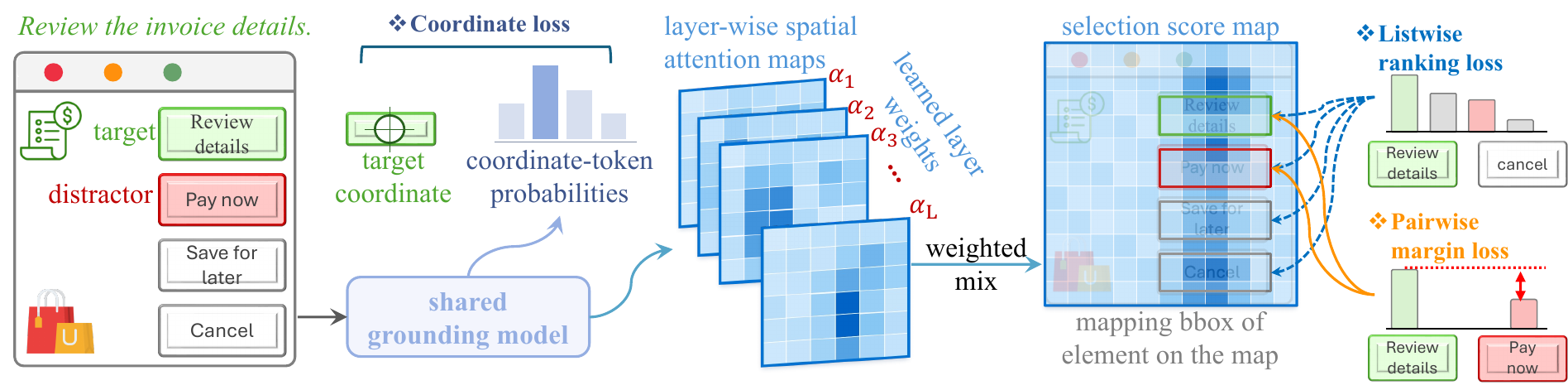}
         \vspace{-5mm}
    \caption{\textbf{Element-wise contrastive training with \modelname{}.} In the example on the left, the target is ``Review details'' and the distractor is ``Pay now.'' Learned weights combine spatial attention maps from the shared grounding model into a selection score map for GUI element regions. The listwise ranking loss contrasts the target with candidate distractors, while the pairwise margin loss encourages a higher score for the target than for the annotated distractor.}
    \label{fig:method_training}
         \vspace{-3mm}
    \end{figure*}

Fig.~\ref{fig:method_training} shows how \modelname{} combines element-wise contrastive learning on observed target--distractor examples with coordinate supervision for GUI grounding. The contrastive objective operates on region selection scores, which are derived from attention and indicate the model preference for each candidate region under the given instruction. We compare the target with distractors in the same screenshot, encouraging a higher score for the target.

\noindent\textbf{Constructing \datasetname{} for contrastive training.} Each standard GUI grounding example contains a screenshot, an instruction $q$, and an annotated target box $\mathrm{bbox}^{+}$, but no distractor annotations. We add a distractor box $\mathrm{bbox}^{-}$ to form target--distractor pairs for contrastive supervision, treating the target as the positive and the distractor as a hard negative. Motivated by the repeated selection of the same non-target elements across models, we use incorrect model predictions on training examples to identify observed distractors. Specifically, we use Click100K as the source of our training data. We obtain observed distractors from incorrect predictions of GTA1-7B and GUI-Libra-8B, matching their click locations to non-target GUI elements detected by OmniParser-v2. This procedure yields 4,292 target--distractor pairs. We combine these pairs with 4,292 ordinary grounding examples sampled across source categories to maintain data diversity, forming \datasetname{}. App.~\ref{sec:app_candidate_verification} provides more details.

\noindent\textbf{Coordinate and contrastive selection losses.} For all examples, we follow standard coordinate-supervised training, using the center of the annotated target box as the target click and applying a token-level cross-entropy loss $\mathcal{L}_{\mathrm{coord}}$ over the target coordinate sequence. For paired examples, we add a contrastive selection loss that encourages a preference for the target over distractor elements. 

Attention-based GUI grounding and visual-score calibration have been explored in prior work~\citep{guiactor,v2p}. Here, we learn a region-selection readout supervised by observed target--distractor pairs. To compare the target with distractor elements, we learn weights to combine attention scores across layers and heads and aggregate them into region selection scores. We use pre-softmax attention scores from the final prompt token to each visual token~\citep{vaswani2017attention}. Let $A_{l,h,i}$ denote this score for visual token $i$ in layer $l$ and head $h$. Learned weights $\alpha_{l,h}$ are normalized by a softmax across layers and heads to produce token selection scores $s_i$, which are aggregated within each region $\mathrm{bbox}$ using log-mean-exp:
\begingroup
\fontsize{8pt}{9pt}\selectfont
\begin{equation}
s_i=\sum_{l,h}\alpha_{l,h}A_{l,h,i},\qquad
S(\mathrm{bbox})=\log\left(\frac{1}{N_{\mathrm{bbox}}}\sum_{i\in\mathrm{bbox}}\exp(s_i)\right).
\label{eq:region_score}
\end{equation}
\endgroup
Here $S(\mathrm{bbox})$ is the score of the entire region, obtained by aggregating its token scores. $N_{\mathrm{bbox}}$ is the number of visual tokens within the box, and $i\in\mathrm{bbox}$ denotes these tokens.

The annotated distractor $\mathrm{bbox}^{-}$ remains fixed for each training example. To introduce additional hard negatives as the model learns, we select up to three other regions with the highest current selection scores, excluding regions whose visual-token masks overlap the target or annotated distractor. The candidate set $\mathcal{B}$ therefore contains $\mathrm{bbox}^{+}$, $\mathrm{bbox}^{-}$, and these additional high-scoring regions. The selection loss $\mathcal{L}_{\mathrm{sel}}$ implements element-wise contrastive supervision. We combine a softmax listwise ranking objective with a margin-augmented pairwise logistic objective, with the pairwise term weighted by $\lambda_{\mathrm{pair}}$:
\par
\begingroup
\fontsize{8pt}{9pt}\selectfont
\begin{equation}
\mathcal{L}_{\mathrm{sel}}=
\underbrace{-\log\frac{\exp(S(\mathrm{bbox}^{+}))}
{\sum_{\mathrm{bbox}\in\mathcal{B}}\exp(S(\mathrm{bbox}))}}_{\mathcal{L}_{\mathrm{list}}\;\text{(Listwise ranking)}}
+\lambda_{\mathrm{pair}}\underbrace{\log\!\left(1+\exp\!\left(
m+S(\mathrm{bbox}^{-})-S(\mathrm{bbox}^{+})\right)\right)}_{\mathcal{L}_{\mathrm{pair}}\;\text{(Pairwise margin)}}.
\label{eq:selection_loss}
\end{equation}  
\endgroup
The listwise ranking loss $\mathcal{L}_{\mathrm{list}}$ contrasts the positive target with all negative distractors in $\mathcal{B}$ through softmax normalization, encouraging a higher score for the target. The pairwise margin loss $\mathcal{L}_{\mathrm{pair}}$ directly compares the target and annotated distractor, using a softplus penalty to encourage the target score to exceed the distractor score by a margin $m$. For paired examples, we add the selection loss weighted by $\lambda_{\mathrm{sel}}$ to the coordinate loss. Ordinary examples use only $\mathcal{L}_{\mathrm{coord}}$. Parsed regions are required only to construct the data and compute the training loss.

\subsection{Region Revisiting and Candidate Comparison: \method{}}

\begin{figure*}[t]
\centering
\includegraphics[width=\textwidth]{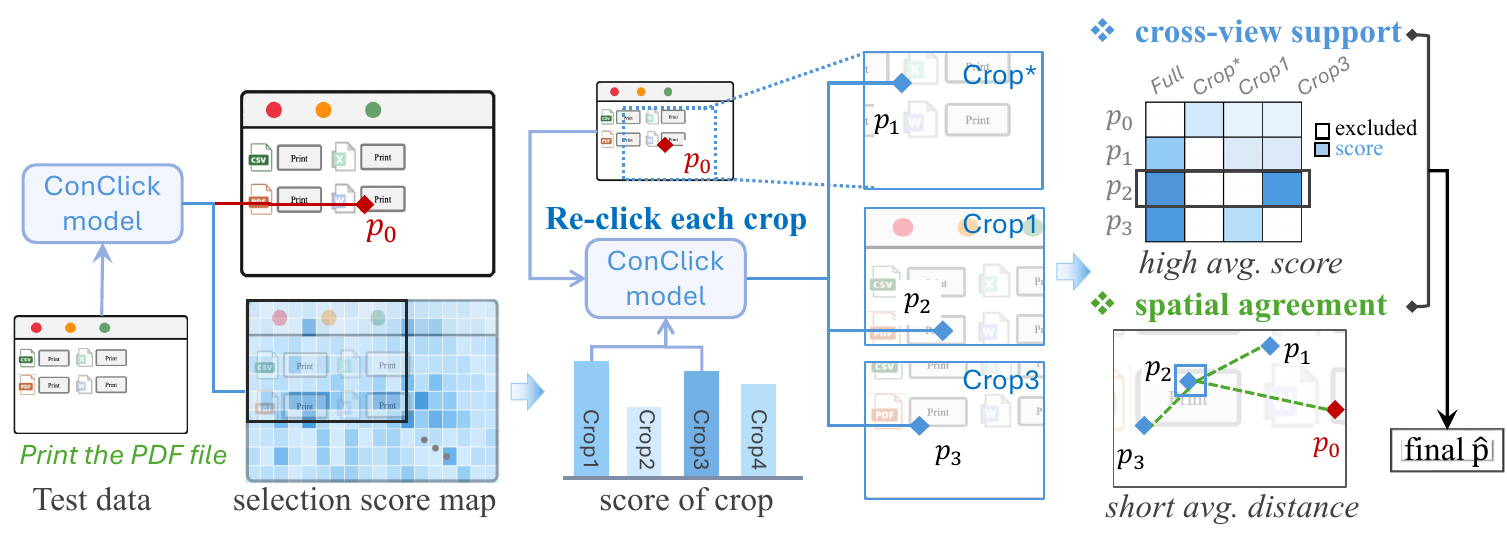}
 \vspace{-5mm}
\caption{\textbf{\method{} at inference time.} Given a screenshot and an instruction, \modelname{} produces an initial click $p_0$ and a selection score map. \method{} selects two high-scoring regions (Crop1 and Crop3) and a crop centered on $p_0$ (Crop*), then predicts clicks $p_1,p_2,p_3$ from the enlarged crops. It compares all four candidates using cross-view support and spatial agreement. Cross-view support measures coordinate likelihood in other views containing the candidate, while spatial agreement favors candidates with smaller mean distances to the others. The final click $\hat p$ maximizes the equally weighted sum of the two standardized scores.}
\label{fig:method_lcr}
 \vspace{-3mm}
\end{figure*} 

Fig.~\ref{fig:method_lcr} illustrates the procedure of \method{}. Region revisits are guided by selection scores learned through contrastive training; for other grounders, these scores can instead be derived from frozen model attention (App.~\ref{sec:app_external_lcr}). Given only a screenshot and an instruction, \method{} first produces an initial prediction, then uses these scores to select regions that may contain competing elements and also examines a crop around the initial click. It compares candidate clicks decoded from these enlarged views and the full screenshot using cross-view likelihood and spatial agreement.

\noindent\textbf{Score regions, revisit crops, and decode clicks.} The full screenshot pass produces the initial prediction $p_0$ and token scores from the attention readout. We define four overlapping crop regions that cover the screenshot, score each region using Eq.~\eqref{eq:region_score}, and select the two highest-scoring regions as potential locations of competing elements (Crop1 and Crop3 in Fig.~\ref{fig:method_lcr}). We also include a crop centered on $p_0$ (Crop* in Fig.~\ref{fig:method_lcr}) to examine the area around the initial click. We enlarge the three crops by $2\times$ and decode one point from each. Mapping these points back to the original screenshot gives candidates $\mathcal P=\{p_0,p_1,p_2,p_3\}$, where $p_1$ comes from the crop around the initial click and $p_2,p_3$ come from the two score-guided crops. The full screenshot and the three enlarged crops form four views for candidate prediction and comparison.

\noindent\textbf{Cross-view candidate evidence.} For each candidate $p$, we compute the log-likelihood of its coordinate sequence in every view containing that location and standardize the scores within each view. We then exclude its generating view and denote the remaining views containing its location by $\mathcal V_p$. For each view $v\in\mathcal V_p$, $\bar\ell_v(p)$ denotes the standardized log-likelihood of the coordinate sequence for $p$ in that view. The cross-view support score is
\par
\begingroup
\fontsize{8pt}{9pt}\selectfont
\begin{equation}
S_{\mathrm{cross}}(p)=\frac{1}{|\mathcal V_p|}\sum_{v\in\mathcal V_p}\bar\ell_v(p).      
\label{eq:lcr_cross} 
\end{equation}
\endgroup
Excluding candidates without cross-view scores gives the remaining candidate set $\mathcal P'$. 

\noindent\textbf{Choosing a candidate.} We measure spatial agreement as the negative mean distance to other candidates in $\mathcal P'$, using coordinates $\tilde p=(p_x/W,p_y/H)$ normalized by screenshot width and height:
\par
\begingroup
\fontsize{8pt}{9pt}\selectfont
\begin{equation}
S_{\mathrm{agree}}(p)=-\frac{\sum_{p'\in\mathcal P'\setminus\{p\}}\|\tilde p-\tilde p'\|_2}{\max(1,|\mathcal P'|-1)}.
\label{eq:lcr_agreement}
\end{equation} 
\endgroup
Higher scores indicate greater spatial agreement; the score is zero when only one candidate remains. We standardize both scores across $\mathcal P'$, denoted by $\hat S_{\mathrm{cross}}$ and $\hat S_{\mathrm{agree}}$, and select the candidate with the highest equally weighted sum:
\par
\begingroup
\fontsize{8pt}{9pt}\selectfont
\begin{equation}
\hat p=\arg\max_{p\in\mathcal P'}\bigl[\hat S_{\mathrm{cross}}(p)+\hat S_{\mathrm{agree}}(p)\bigr]. 
\label{eq:lcr_select}
\end{equation}
\endgroup
We use the same comparison rule across all three benchmarks and retain the initial prediction if $\mathcal P'$ is empty. The procedure uses up to four visual grounding passes plus cached-prefix candidate continuations. App.~\ref{sec:app_inference_details} provides more implementation details.

\section{Experiments}
\subsection{Experimental Setups}
\label{sec:experiments_setup}

\noindent\textbf{Training data and backbone.} Section~\ref{sec:training_observed} describes the training data. Our primary backbone is Qwen3-VL-8B. We also evaluate InternVL3.5-8B and Qwen3-VL-30B-A3B-Instruct to verify cross-backbone applicability. We use LoRA~\citep{lora} with bfloat16 for every backbone. App.~\ref{sec:app_semantic_data_details} provides the detailed training settings for \modelname{}-8B.

\noindent\textbf{Evaluation.} Following prior work, we report accuracy (Acc.), the percentage of predictions inside the target region. We also report the element error rate (E-err.), the percentage of all predictions that fall outside the target region and on a parsed GUI element. We evaluate on ScreenSpot-Pro, MMBench-GUI L2, and OSWorld-G, with 1,581, 3,594, and 510 examples, respectively.

\noindent\textbf{Inference baselines.} We compare \method{} with MVP~\citep{mvp} and ZoomClick~\citep{zoomclick} on the same \modelname{}-8B checkpoint. MVP uses its released prompts and preprocessing, while ZoomClick uses its released search procedure. App.~\ref{sec:app_inference_details} provides implementation details, including minimum crop sizes and boundary handling.

\begin{table*}[t]
\centering 
\fontsize{8pt}{9pt}\selectfont
\setlength{\tabcolsep}{4.3pt}   
\caption{Performance of different models.
Acc. is point accuracy, and E-err. is the element error rate over all predictions.
\modelname{}-8B and \modelname{}-8B + \method{} use the same trained checkpoint.
For other models, $^{*}$ denotes published accuracies reported in the original publications; all other results are re-evaluated using released weights and implementations.
\srcverified{} marks ScreenSpot-Pro accuracy within one percentage point of the published result, supporting the fidelity of the reproduction used to compute E-err.
Avg. is the mean across the three benchmarks.
All values are percentages. \textcolor{blue}{\textbf{Best}} and \textbf{second-best} results are in \textcolor{blue}{\textbf{blue bold}} and \textbf{black bold}, respectively.}
\label{tab:main_results}
\begin{tabular}{lll|ll|ll|ll} 
\toprule
& \multicolumn{2}{c}{ScreenSpot-Pro} & \multicolumn{2}{c}{OSWorld-G}  & \multicolumn{2}{c}{MMB-GUI L2} & \multicolumn{2}{c}{Avg.} \\
\cmidrule(lr){2-3}\cmidrule(lr){4-5}\cmidrule(lr){6-7}\cmidrule(lr){8-9}
\multicolumn{1}{l}{Method} & \multicolumn{1}{l}{Acc.$\uparrow$} & \multicolumn{1}{l}{E-err.$\downarrow$} & \multicolumn{1}{l}{Acc.$\uparrow$} & \multicolumn{1}{l}{E-err.$\downarrow$} & \multicolumn{1}{l}{Acc.$\uparrow$} & \multicolumn{1}{l}{E-err.$\downarrow$} & \multicolumn{1}{l}{Acc.$\uparrow$} & \multicolumn{1}{l}{E-err.$\downarrow$} \\
\midrule
GUI-Actor-7B~\citep{guiactor} & 44.09\srcverified & 43.83 & 43.92 & 42.35  & 69.59 & 19.37 & 52.53 & 35.18 \\
SE-GUI-7B~\citep{segui} & 46.74\srcverified & 37.51 & 49.02 & 42.16  & 76.74 & 19.89 & 57.50 & 33.19 \\
GUI-G2-7B~\citep{guig2} & 48.39\srcverified & 33.14 & 40.59 & 41.37  & 79.27 & 17.84 & 56.08 & 30.78 \\
OpenCUA-7B~\citep{opencua} & 49.46\srcverified & 34.98 & 60.98 & 31.96  & 81.02 & 16.17 & 63.82 & 27.70 \\
UI-TARS-1.5-7B~\citep{uitars15} & 49.65\srcverified & 33.02 & 60.78 & 31.76  & 75.43 & 20.67 & 61.95 & 28.48 \\
V2P-7B~\citep{v2p} & 52.25\srcverified & 41.68 & 58.24 & 36.08  & 80.30 & 18.42 & 63.60 & 32.06 \\
GUI-Libra-8B~\citep{guilibra} & 54.84\srcverified & 38.20 & 63.53 & 33.33  & 79.60 & 19.09 & 65.99 & 30.21 \\
GTA1-7B~\citep{gta1} & 55.34\srcverified & 35.36 & 64.71 & 29.22  & 83.86 & 14.83 & 67.97 & 26.47 \\
\midrule
ZwZ-8B~\citep{zwz2026} & 56.80\textsuperscript{*} & - & - & -  & - & - & - & - \\
Propose-then-Critic-8B~\citep{measuretwice} & 58.70\textsuperscript{*} & - & - & -  & 78.40\textsuperscript{*} & - & - & - \\
\midrule
Qwen3-VL-8B~\citep{qwen3vl} (baseline) & 54.71 & 33.84 & 63.33 & 29.22  & 83.72 & 14.66 & 67.25 & 25.91 \\
\modelname{}-8B (Ours) &  \textbf{66.03} & \textbf{30.30} & \textbf{70.20} & \textbf{24.12}  & \textbf{86.28} & \textbf{13.08} & \textbf{74.17} & \textbf{22.50} \\
\modelname{}-8B + \method{} (Ours)  & \textcolor{blue}{\textbf{72.36}} & \textcolor{blue}{\textbf{25.87}} & \textcolor{blue}{\textbf{71.76}} & \textcolor{blue}{\textbf{23.73}}  & \textcolor{blue}{\textbf{87.28}} & \textcolor{blue}{\textbf{12.10}} & \textcolor{blue}{\textbf{77.13}} & \textcolor{blue}{\textbf{20.57}} \\
\bottomrule
\end{tabular}
\end{table*}

\subsection{Main Results} 

\noindent\textbf{GUI Model Comparisons.} Table~\ref{tab:main_results} shows that \modelname{}-8B improves accuracy and reduces element error rates over the Qwen3-VL-8B backbone on all three benchmarks. It achieves accuracies of 66.03\%, 86.28\%, and 70.20\% on ScreenSpot-Pro, MMBench-GUI L2, and OSWorld-G, respectively, improving over the baseline by 11.32, 2.56, and 6.87 percentage points. Compared with other locally evaluated models in Table~\ref{tab:main_results}, \modelname{}-8B achieves the highest average accuracy (74.17\%) and the lowest element error rates on all three benchmarks, averaging 22.50\%; error-rate comparisons are limited to methods with available results. These results support our training strategy for improving grounding accuracy and reducing element selection errors. Subset accuracies for locally evaluated models appear in App. Tables~\ref{tab:app_ssp_subsets} (ScreenSpot-Pro) and~\ref{tab:app_mmb_subsets} (MMBench-GUI L2).

\noindent\textbf{\method{} further improves grounding at inference time.} Compared with direct inference, \method{} increases accuracy by 6.33, 1.00, and 1.57 percentage points on these three datasets, respectively, while reducing element error rates on all three benchmarks (Table~\ref{tab:main_results}).

\noindent\textbf{Higher grounding accuracy does not necessarily imply fewer clicks on wrong elements.} Across the nine prior models with both metrics in Table~\ref{tab:main_results}, the Spearman correlations between accuracy and element error rate are $-0.117$, $-0.883$, and $-0.820$ on ScreenSpot-Pro, MMBench-GUI L2, and OSWorld-G, respectively. Although the two metrics are negatively correlated on the latter two benchmarks, their weak correlation on ScreenSpot-Pro shows that accuracy alone does not reliably reflect element selection errors across models. For example, GUI-Libra-8B has higher accuracy than UI-TARS-1.5-7B on ScreenSpot-Pro (54.84\% vs.\ 49.65\%) but also a higher element error rate (38.20\% vs.\ 33.02\%). We therefore recommend reporting error types, particularly element errors, alongside accuracy when evaluating GUI grounding models.

\begin{figure*}[t]
    \centering
    \includegraphics[width=\textwidth]{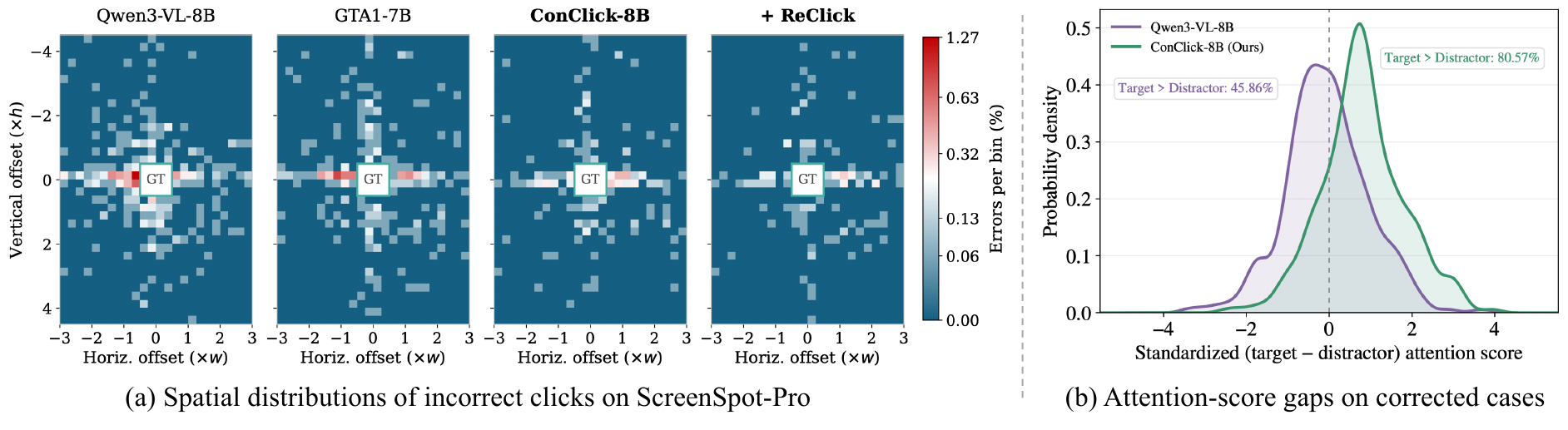}
     \vspace{-4mm}
    \caption{\textbf{Spatial errors and attention-score gaps.} (a) Incorrect clicks on ScreenSpot-Pro, with target-centered offsets normalized by target width and height. Red indicates more errors per bin as a percentage of all predictions. (b) Standardized target--distractor score gaps on Click100K Val examples where Qwen3-VL-8B fails and \modelname{}-8B succeeds.}
    \label{fig:near_error_density}
     \vspace{-4mm}
\end{figure*}

\noindent\textbf{\modelname{} and \method{} also reduce near-target error density.} Fig.~\ref{fig:near_error_density}(a) compares the spatial distributions of incorrect clicks on ScreenSpot-Pro. Compared with the Qwen3-VL-8B baseline, our training strategy lowers error density within the target neighborhood, and \method{} further reduces it at inference time. \modelname{}-8B exhibits lower near-target error density than GTA1-7B. These results show that improvements extend beyond element selection to reducing errors near the target.

\noindent\textbf{Attention scores favor the target on corrected examples.} Fig.~\ref{fig:near_error_density}(b) examines examples from Click100K Val on which Qwen3-VL-8B predicts incorrectly but \modelname{}-8B predicts correctly. We compute the attention-score difference between the target and a fixed distractor for each model, standardized by the standard deviation of its visual-token scores. Compared with Qwen3-VL-8B, the distribution for \modelname{}-8B shifts toward more positive values, indicating a stronger preference for the target over the distractor on these corrected examples. These results support the effectiveness of \modelname{} in strengthening the preference for the target over distractors.

\begin{table*}[t]
    \centering
    \fontsize{8pt}{9pt}\selectfont
    \setlength{\tabcolsep}{4.3pt}
    \caption{Ablations of training data and loss components on Qwen3-VL-8B.
    Colors are normalized within each column, with darker green indicating better performance; bold marks the best results.}
    \label{tab:training_inference_transfer}
    \begin{tabular}{c@{\hspace{6pt}}c@{\hspace{10pt}}c@{\hspace{6pt}}c@{\hspace{6pt}}crr|rr|rr|rr}
    \toprule
    \multicolumn{2}{c}{Data}
    & \multicolumn{3}{c}{Loss}
    & \multicolumn{2}{c}{ScreenSpot-Pro}
    & \multicolumn{2}{c}{OSWorld-G}
    & \multicolumn{2}{c}{MMB-GUI L2}
    & \multicolumn{2}{c}{Avg.} \\
    \cmidrule(lr){1-2}\cmidrule(lr){3-5}
    \cmidrule(lr){6-7}\cmidrule(lr){8-9}
    \cmidrule(lr){10-11}\cmidrule(lr){12-13}

    Random & Our
    & $\mathcal{L}_{\mathrm{coord}}$
    & $\mathcal{L}_{\mathrm{list}}$
    & $\mathcal{L}_{\mathrm{pair}}$
    & \multicolumn{1}{c}{Acc.$\uparrow$}
    & \multicolumn{1}{c}{E-err.$\downarrow$}
    & \multicolumn{1}{c}{Acc.$\uparrow$}
    & \multicolumn{1}{c}{E-err.$\downarrow$}
    & \multicolumn{1}{c}{Acc.$\uparrow$}
    & \multicolumn{1}{c}{E-err.$\downarrow$}
    & \multicolumn{1}{c}{Acc.$\uparrow$}
    & \multicolumn{1}{c}{E-err.$\downarrow$} \\
    \midrule

    & & & & &
    \cellcolor[HTML]{F2F8F4}54.71 &
    \cellcolor[HTML]{F2F8F4}33.84 &
    \cellcolor[HTML]{F2F8F4}63.33 &
    \cellcolor[HTML]{F2F8F4}29.22 &
    \cellcolor[HTML]{F2F8F4}83.72 &
    \cellcolor[HTML]{F2F8F4}14.66 &
    \cellcolor[HTML]{F2F8F4}67.25 &
    \cellcolor[HTML]{F2F8F4}25.91 \\
    $\checkmark$ & & $\checkmark$ & & &
    \cellcolor[HTML]{E9F3EC}59.90 &
    \cellcolor[HTML]{ECF5EF}32.64 &
    \cellcolor[HTML]{D4E8DB}67.65 &
    \cellcolor[HTML]{ECF5EF}26.08 &
    \cellcolor[HTML]{D7EADE}84.47 &
    \cellcolor[HTML]{D3E8DA}14.16 &
    \cellcolor[HTML]{E0EFE5}70.67 &
    \cellcolor[HTML]{E8F3EC}24.29 \\
    
& $\checkmark$ & $\checkmark$ & & &
\cellcolor[HTML]{CAE3D3}62.43 &
\cellcolor[HTML]{D2E7D9}32.26 &
\cellcolor[HTML]{BFDECA}68.63 &
\cellcolor[HTML]{DEEEE4}25.49 &
\cellcolor[HTML]{C8E2D1}84.92 &
\cellcolor[HTML]{BFDECA}13.72 &
\cellcolor[HTML]{C6E1D0}71.99 &
\cellcolor[HTML]{D5E9DC}23.82 \\

    & $\checkmark$ & $\checkmark$ & $\checkmark$ & &
    \cellcolor[HTML]{A6D0B6}64.26 &
    \cellcolor[HTML]{BFDDCA}31.82 &
    \cellcolor[HTML]{98C9AA}\textbf{70.20} &
    \cellcolor[HTML]{A9D2B8}25.10 &
    \cellcolor[HTML]{A9D2B8}85.81 &
    \cellcolor[HTML]{ABD3B9}13.41 &
    \cellcolor[HTML]{A2CEB2}73.42 &
    \cellcolor[HTML]{B1D6BE}23.44 \\

    & $\checkmark$ & $\checkmark$ & & $\checkmark$ &
    \cellcolor[HTML]{A7D1B6}64.20 &
    \cellcolor[HTML]{C5E0CF}32.07 &
    \cellcolor[HTML]{A2CEB3}69.41 &
    \cellcolor[HTML]{ADD4BB}25.29 &
    \cellcolor[HTML]{A5D0B4}85.92 &
    \cellcolor[HTML]{A8D1B7}13.36 &
    \cellcolor[HTML]{A5D0B5}73.18 &
    \cellcolor[HTML]{B4D8C1}23.57 \\

    & $\checkmark$ & $\checkmark$ & $\checkmark$ & $\checkmark$ &
    \cellcolor[HTML]{98C9AA}\textbf{66.03} &
    \cellcolor[HTML]{98C9AA}\textbf{30.30} &
    \cellcolor[HTML]{98C9AA}\textbf{70.20} &
    \cellcolor[HTML]{98C9AA}\textbf{24.12} &
    \cellcolor[HTML]{98C9AA}\textbf{86.28} &
    \cellcolor[HTML]{98C9AA}\textbf{13.08} &
    \cellcolor[HTML]{98C9AA}\textbf{74.17} &
    \cellcolor[HTML]{98C9AA}\textbf{22.50} \\

    \bottomrule
    \end{tabular}
\end{table*}

\subsection{Ablation studies} 
\label{sec:ablations}

\noindent\textbf{Ablations for \modelname{}.} Table~\ref{tab:training_inference_transfer} examines the contributions of training data and contrastive losses. With coordinate-only training, replacing equally sized random data with our data raises average accuracy from 70.67\% to 71.99\% and lowers the average element error rate from 24.29\% to 23.82\%. Adding the full contrastive objective further improves these metrics to 74.17\% and 22.50\%, respectively. On the same data, combining the listwise ranking loss $\mathcal{L}_{\mathrm{list}}$ and pairwise margin loss $\mathcal{L}_{\mathrm{pair}}$ (Eq.~\eqref{eq:selection_loss}) achieves higher average accuracy than either term alone (74.17\% vs.\ 73.42\% and 73.18\%, respectively), with the lowest element error rates across all three benchmarks.   These results support the effectiveness of both our data construction strategy and the combined contrastive objective. App.~\ref{sec:app_candidate_verification} details the random-data sampling procedure, and App.~\ref{sec:app_additional_backbones} reports additional training ablations on InternVL3.5-8B and Qwen3-VL-30B-A3B-Instruct.

\begin{figure*}[t]
    \centering
    \includegraphics[width=\textwidth]{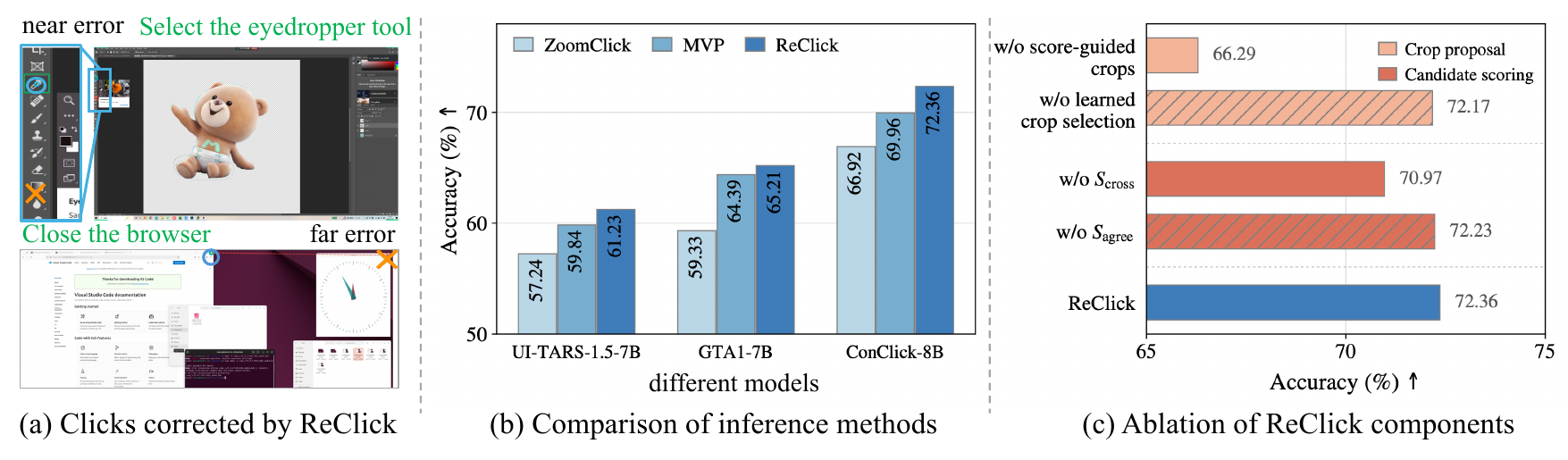}
     \vspace{-4mm}
    \caption{\textbf{Examples, comparisons, and ablations of \method{}.} (a) Examples where \method{} corrects initial clicks near and far from the target. Orange crosses mark initial incorrect clicks, and blue circles mark corrected predictions. (b) Accuracy comparison of ZoomClick, MVP, and \method{} across three backbones. (c) Ablations of crop proposal and candidate scoring components using the same \modelname{}-8B checkpoint. Results in (b,c) are on ScreenSpot-Pro.
    }
    \label{fig:lcr_analysis}
     \vspace{-3mm}
\end{figure*}

\noindent\textbf{Test-time method comparisons and \method{} ablations.} Fig.~\ref{fig:lcr_analysis}(a) shows \method{} correcting initial clicks both near and far from the target, illustrating its benefits extend beyond nearby errors. 
Fig.~\ref{fig:lcr_analysis}(b) shows \method{} achieving higher accuracy than ZoomClick and MVP on ScreenSpot-Pro across three GUI grounding models. On \modelname{}-8B, it reaches 72.36\% accuracy, compared with 69.96\% for MVP and 66.92\% for ZoomClick. Applying \method{} to other backbones requires adapting the attention readout. Implementation details appear in App.~\ref{sec:app_external_lcr}.
Fig.~\ref{fig:lcr_analysis}(c) examines the components of \method{} on ScreenSpot-Pro. Removing score-guided crops or using a fixed attention readout to select them lowers average accuracy across the three benchmarks. Combining cross-view support and spatial agreement yields the best average accuracy (App. Table~\ref{tab:lcr_ablation}).

\section{Related Work}
\noindent\textbf{GUI grounding training.} Prior work develops coordinate pretraining and grounding supervision across platforms~\citep{seeclick,uground,osatlas,osworldg,groundcua}, attention based action prediction and visual scoring~\citep{guiactor,v2p}, and reinforcement learning or spatial and semantic alignment~\citep{guig1,segui,guig2,infiguig1}. VISTA and UI-Ins explore target preserving views and multi-perspective instructions, respectively~\citep{vista,uiins}. Related studies examine instruction sensitivity and hallucinations~\citep{doguiunderstand,guihallubench}. Motivated by our error analysis, \modelname{} contrasts target and observed distractor region scores alongside coordinate supervision to improve element selection.

\noindent\textbf{Inference-time improvements.} Prior methods use region search, cropping, and candidate selection~\citep{screenspotpro,regionfocus,uizoomer,zoomclick,localizationbias,bami}, or reprocess selected visual tokens~\citep{focusui,reprefill}. MVP clusters predictions across attention guided views~\citep{mvp}, while GUI-RC uses spatial voting over sampled predictions~\citep{guirc}. UI-AGILE uses a vision language model to select crops, and Measure Twice trains a visual critic~\citep{uiagile,measuretwice}. Our \method{} requires no external model beyond the trained grounder; it revisits selected regions and combines cross-view likelihood with spatial agreement to select the final click.

\section{Conclusion}
This paper analyzes GUI grounding errors across ten models and five datasets, showing that most incorrect clicks hit non-target elements and different models repeatedly select the same distractors. Motivated by these findings, we construct \datasetname{} and introduce \modelname{}, an element-wise contrastive training framework that learns to distinguish targets from observed distractors. As a natural extension, \method{} uses learned region preferences to guide revisits and compares candidate predictions across views. Experiments show improved grounding accuracy and fewer clicks on non-target elements, with further gains from \method{}. Our findings highlight the importance of distinguishing error types alongside accuracy, since wrong-element clicks can trigger harmful actions, and show how error analysis can guide training and inference.

\bibliography{references}

@article{qwen3vl,
  title={Qwen3-vl technical report},
  author={Bai, Shuai and Cai, Yuxuan and Chen, Ruizhe and Chen, Keqin and Chen, Xionghui and Cheng, Zesen and Deng, Lianghao and Ding, Wei and Gao, Chang and Ge, Chunjiang and others},
  journal={arXiv preprint arXiv:2511.21631},
  year={2025}
}

@article{internvl35,
  title={Internvl3. 5: Advancing open-source multimodal models in versatility, reasoning, and efficiency},
  author={Wang, Weiyun and Gao, Zhangwei and Gu, Lixin and Pu, Hengjun and Cui, Long and Wei, Xingguang and Liu, Zhaoyang and Jing, Linglin and Ye, Shenglong and Shao, Jie and others},
  journal={arXiv preprint arXiv:2508.18265},
  year={2025}
}

@inproceedings{osatlas,
  title={OS-ATLAS: Foundation action model for generalist GUI agents},
  author={Wu, Zhiyong and Wu, Zhenyu and Xu, Fangzhi and Wang, Yian and Sun, Qiushi and Jia, Chengyou and Cheng, Kanzhi and Ding, Zichen and Chen, Liheng and Liang, Paul Pu and others},
  booktitle={International Conference on Learning Representations},
  volume={2025},
  pages={5090--5108},
  year={2025}
}

@article{uitars,
  title={Ui-tars: Pioneering automated gui interaction with native agents},
  author={Qin, Yujia and Ye, Yining and Fang, Junjie and Wang, Haoming and Liang, Shihao and Tian, Shizuo and Zhang, Junda and Li, Jiahao and Li, Yunxin and Huang, Shijue and others},
  journal={arXiv preprint arXiv:2501.12326},
  year={2025}
}

@article{osworldg,
  title={Scaling computer-use grounding via user interface decomposition and synthesis},
  author={Xie, Tianbao and Deng, Jiaqi and Li, Xiaochuan and Yang, Junlin and Wu, Haoyuan and Chen, Jixuan and Hu, Wenjing and Wang, Xinyuan and Xu, Yuhui and Wang, Zekun and others},
  journal={Advances in Neural Information Processing Systems},
  volume={38},
  year={2026}
}

@article{uivenus,
  title={Ui-venus technical report: Building high-performance ui agents with rft},
  author={Gu, Zhangxuan and Zeng, Zhengwen and Xu, Zhenyu and Zhou, Xingran and Shen, Shuheng and Liu, Yunfei and Zhou, Beitong and Meng, Changhua and Xia, Tianyu and Chen, Weizhi and others},
  journal={arXiv preprint arXiv:2508.10833},
  year={2025}
}

@article{segui,
  title={Enhancing visual grounding for gui agents via self-evolutionary reinforcement learning},
  author={Yuan, Xinbin and Zhang, Jian and Li, Kaixin and Cai, Zhuoxuan and Yao, Lujian and Chen, Jie and Wang, Enguang and Hou, Qibin and Chen, Jinwei and Jiang, Peng-Tao and others},
  journal={arXiv preprint arXiv:2505.12370},
  year={2025}
}

@article{guig1,
  title={Gui-g1: Understanding r1-zero-like training for visual grounding in gui agents},
  author={Zhou, Yuqi and Dai, Sunhao and Wang, Shuai and Zhou, Kaiwen and Jia, Qinglin and Xu, Jun},
  journal={Advances in Neural Information Processing Systems},
  volume={38},
  pages={95683--95705},
  year={2026}
}

@inproceedings{gta1,
  title={Gta1: Gui test-time scaling agent},
  author={Yang, Yan and Li, Dongxu and Dai, Yutong and Yang, Yuhao and Luo, Ziyang and Zhao, Zirui and Hu, Zhiyuan and Huang, Junzhe and Saha, Amrita and Chen, Zeyuan and others},
  booktitle={International Conference on Learning Representations},
  volume={2026},
  pages={48295--48323},
  year={2026}
}

@inproceedings{screenspotpro,
  title={Screenspot-pro: Gui grounding for professional high-resolution computer use},
  author={Li, Kaixin and Meng, Ziyang and Lin, Hongzhan and Luo, Ziyang and Tian, Yuchen and Ma, Jing and Huang, Zhiyong and Chua, Tat-Seng},
  booktitle={Proceedings of the 33rd ACM International Conference on Multimedia},
  pages={8778--8786},
  year={2025}
}

@inproceedings{mvp,
  title={Mvp: Multiple view prediction improves gui grounding},
  author={Zhang, Yunzhu and Pan, Zeyu and Zeng, Zhengwen and Shen, Shuheng and Meng, Changhua and Zhu, Linchao},
  booktitle={Proceedings of the IEEE/CVF Conference on Computer Vision and Pattern Recognition},
  pages={27482--27492},
  year={2026}
}

@article{uivision,
  title={Ui-vision: A desktop-centric gui benchmark for visual perception and interaction},
  author={Nayak, Shravan and Jian, Xiangru and Lin, Kevin Qinghong and Rodriguez, Juan A and Kalsi, Montek and Awal, Rabiul and Chapados, Nicolas and {\"O}zsu, M Tamer and Agrawal, Aishwarya and Vazquez, David and others},
  journal={arXiv preprint arXiv:2503.15661},
  year={2025}
}

@article{mmbenchgui,
  title={Mmbench-gui: Hierarchical multi-platform evaluation framework for gui agents},
  author={Wang, Xuehui and Wu, Zhenyu and Xie, JingJing and Ding, Zichen and Yang, Bowen and Li, Zehao and Liu, Zhaoyang and Li, Qingyun and Dong, Xuan and Chen, Zhe and others},
  journal={arXiv preprint arXiv:2507.19478},
  year={2025}
}

@article{opencua,
  title={Opencua: Open foundations for computer-use agents},
  author={Wang, Xinyuan and Wang, Bowen and Lu, Dunjie and Yang, Junlin and Xie, Tianbao and Wang, Junli and Deng, Jiaqi and Guo, Xiaole and Xu, Yiheng and Wu, Chen and others},
  journal={Advances in Neural Information Processing Systems},
  volume={38},
  pages={139756--139806},
  year={2026}
}

@article{v2p,
  title={V2P: Visual Attention Calibration for GUI Grounding via Background Suppression and Center Peaking},
  author={Chen, Jikai and Chen, Long and Wang, Dong and Su, Qinglin and Chu, Zhixuan and Hao, Bingguang and Gan, Leilei and Zhuang, Chenyi and Gu, Jinjie},
  journal={arXiv preprint arXiv:2508.13634},
  year={2025}
}

@inproceedings{uiins,
  title={UI-Ins: Enhancing GUI grounding with multi-perspective instruction as reasoning},
  author={Chen, Liangyu and Zhou, Hanzhang and Zhang, Jianan and Tong, Panrong and Zhang, Xu and Kong, Quyu and Liu, Chen and Liu, Yuqi and Wang, Wenxuan and Wang, Yue and others},
  booktitle={International Conference on Learning Representations},
  volume={2026},
  pages={54418--54444},
  year={2026}
}

@article{guiactor,
  title={Gui-actor: Coordinate-free visual grounding for gui agents},
  author={Wu, Qianhui and Cheng, Kanzhi and Yang, Rui and Zhang, Chaoyun and Yang, Jianwei and Jiang, Huiqiang and Mu, Jian and Peng, Baolin and Qiao, Bo and Tan, Reuben and others},
  journal={Advances in Neural Information Processing Systems},
  volume={38},
  pages={15101--15128},
  year={2026}
}

@misc{uitars15,
  title = {UI-TARS-1.5},
  author = {ByteDance Seed},
  howpublished = {\url{https://seed-tars.com/1.5}},
  year = {2025},
}

@inproceedings{guig2,
  title={GUI-G$^2$: Gaussian Reward Modeling for GUI Grounding},
  author={Tang, Fei and Gu, Zhangxuan and Lu, Zhengxi and Liu, Xuyang and Shen, Shuheng and Meng, Changhua and Wang, Wen and Zhang, Wenqi and Shen, Yongliang and Lu, Weiming and others},
  booktitle={Proceedings of the AAAI Conference on Artificial Intelligence},
  volume={40},
  number={39},
  pages={33214--33222},
  year={2026}
}

@article{omniparserv2,
  title={OmniParser V2: Turning any LLM into a computer use agent},
  author={Lu, Y and Dhome-Casanova, T and Yang, J and Awadallah, A},
  journal={Microsoft Research. Retrieved March},
  volume={18},
  pages={2025},
  year={2025}
}

@article{zoomclick,
  title={Zoom in, click out: Unlocking and evaluating the potential of zooming for gui grounding},
  author={Jiang, Zhiyuan and Xie, Shenghao and Li, Wenyi and Zu, Wenqiang and Li, Peihang and Qiu, Jiahao and Pei, Siqi and Ma, Lei and Huang, Tiejun and Wang, Mengdi and others},
  journal={arXiv preprint arXiv:2512.05941},
  year={2025}
}

@article{uizoomer,
  title={Ui-zoomer: Uncertainty-driven adaptive zoom-in for gui grounding},
  author={Tang, Fei and Chen, Bofan and Lu, Zhengxi and Chen, Tongbo and Nong, Songqin and Jiang, Tao and Xu, Wenhao and Lu, Weiming and Xiao, Jun and Zhuang, Yueting and others},
  journal={arXiv preprint arXiv:2604.14113},
  year={2026}
}

@article{reprefill,
  title={What Happens Before Decoding? Prefill Determines GUI Grounding in VLMs},
  author={Lin, Jiaping and Shen, Fei and Li, Junzhe and Nie, Ping and Yu, Fei and Li, Ming and Li, Haizhou},
  journal={arXiv preprint arXiv:2605.12549},
  year={2026}
}

@article{focusui,
  title={Focusui: Efficient ui grounding via position-preserving visual token selection},
  author={Ouyang, Mingyu and Lin, Kevin Qinghong and Shou, Mike Zheng and Ng, Hwee Tou},
  journal={arXiv preprint arXiv:2601.03928},
  year={2026}
}

@article{vista,
  title={VISTA: View-Consistent Self-Verified Training for GUI Grounding},
  author={Qiu, Xinyu and Zhang, Yunzhu and Jia, Heng and Shen, Shuheng and Meng, Changhua and Zhu, Linchao},
  journal={arXiv preprint arXiv:2606.14579},
  year={2026}
}

@inproceedings{seeclick,
  title={Seeclick: Harnessing gui grounding for advanced visual gui agents},
  author={Cheng, Kanzhi and Sun, Qiushi and Chu, Yougang and Xu, Fangzhi and YanTao, Li and Zhang, Jianbing and Wu, Zhiyong},
  booktitle={Proceedings of the 62nd Annual Meeting of the Association for Computational Linguistics (Volume 1: Long Papers)},
  pages={9313--9332},
  year={2024}
}

@inproceedings{uground,
  title={Navigating the digital world as humans do: Universal visual grounding for gui agents},
  author={Gou, Boyu and Wang, Demi Ruohan and Zheng, Boyuan and Xie, Yanan and Chang, Cheng and Shu, Yiheng and Sun, Huan and Su, Yu},
  booktitle={International Conference on Learning Representations},
  volume={2025},
  pages={30851--30883},
  year={2025}
}

@article{bami,
  title={BAMI: Training-Free Bias Mitigation in GUI Grounding},
  author={Zhang, Borui and Zhang, Bo and Wang, Bo and Zheng, Wenzhao and Cheng, Yuhao and Tang, Liang and Yan, Yiqiang and Zhou, Jie and Lu, Jiwen},
  journal={arXiv preprint arXiv:2605.06664},
  year={2026}
}

@inproceedings{guihallubench,
  title={Exposing and Evaluating Hallucinations for GUI Grounding},
  author={Zhang, Zicheng and Jing, Hongyi and Lv, Rui and Fang, Shuo and Zhu, Shiai and Wang, Junying and Li, Chunyi and Liu, Xiaohong and Ma, Chenguang and Zhai, Guangtao},
  booktitle={Proceedings of the IEEE/CVF Conference on Computer Vision and Pattern Recognition},
  pages={40208--40223},
  year={2026}
}

@inproceedings{doguiunderstand,
  title={Do GUI Grounders Truly Understand UI Elements?},
  author={Jandial, Surgan and Li, Yinheng and Wagle, Justin and Koishida, Kazuhito},
  booktitle={Findings of the Association for Computational Linguistics: EACL 2026},
  pages={2772--2785},
  year={2026}
}

@article{vaswani2017attention,
  title={Attention is all you need},
  author={Vaswani, Ashish and Shazeer, Noam and Parmar, Niki and Uszkoreit, Jakob and Jones, Llion and Gomez, Aidan N and Kaiser, {\L}ukasz and Polosukhin, Illia},
  journal={Advances in neural information processing systems},
  volume={30},
  year={2017}
}

@inproceedings{guirc,
  title={Test-time reinforcement learning for gui grounding via region consistency},
  author={Du, Yong and Yan, Yuchen and Tang, Fei and Lu, Zhengxi and Zong, Chang and Lu, Weiming and Jiang, Shengpei and Shen, Yongliang},
  booktitle={Proceedings of the AAAI Conference on Artificial Intelligence},
  volume={40},
  number={36},
  pages={30593--30601},
  year={2026}
}

@inproceedings{groundcua,
  title={Grounding computer use agents on human demonstrations},
  author={Feizi, Aarash and Nayak, Shravan and Jian, Xiangru and Lin, Kevin Qinghong and Li, Kaixin and Awal, Rabiul and L{\`u}, Xing Han and Obando Ceron, Johan S and Rodriguez, Juan A and Chapados, Nicolas and others},
  booktitle={International Conference on Learning Representations},
  volume={2026},
  pages={5181--5215},
  year={2026}
}

@inproceedings{infiguig1,
  title={Infigui-g1: Advancing gui grounding with adaptive exploration policy optimization},
  author={Liu, Yuhang and Liu, Zeyu and Zhu, Shuanghe and Li, Pengxiang and Xie, Congkai and Wang, Jiasheng and Hu, Xueyu and Han, Xiaotian and Yuan, Jianbo and Wang, Xinyao and others},
  booktitle={Proceedings of the AAAI Conference on Artificial Intelligence},
  volume={40},
  number={38},
  pages={32267--32275},
  year={2026}
}

@inproceedings{localizationbias,
  title={Understanding gui agent localization biases through logit sharpness},
  author={Tao, Xingjian and Wang, Yiwei and Cai, Yujun and Yang, Zhicheng and Tang, Jing},
  booktitle={Findings of the Association for Computational Linguistics: EMNLP 2025},
  pages={23361--23374},
  year={2025}
}

@inproceedings{uiagile,
  title={Ui-agile: Advancing gui agents with effective reinforcement learning and precise inference-time grounding},
  author={Lian, Shuquan and Wu, Yuhang and Ma, Jia and Ding, Yifan and Song, Zihan and Chen, Bingqi and Zheng, Xiawu and Li, Hui and Ji, Rongrong},
  booktitle={Proceedings of the IEEE/CVF Conference on Computer Vision and Pattern Recognition},
  pages={8575--8584},
  year={2026}
}

@inproceedings{measuretwice,
  title={Measure Twice, Click Once: Co-evolving Proposer and Visual Critic via Reinforcement Learning for GUI Grounding},
  author={Wang, Wenkai and Li, Xiyun and Guo, Hongcan and Yu, Wenhao and Fang, Tianqing and Mi, Haitao and Yu, Dong and Zhang, Shengyu},
  booktitle={Proceedings of the 64th Annual Meeting of the Association for Computational Linguistics (Volume 1: Long Papers)},
  pages={20964--20984},
  year={2026}
}

@inproceedings{lora,
title={Lo{RA}: Low-Rank Adaptation of Large Language Models},
author={Edward J Hu and Yelong Shen and Phillip Wallis and Zeyuan Allen-Zhu and Yuanzhi Li and Shean Wang and Lu Wang and Weizhu Chen},
booktitle={International Conference on Learning Representations},
year={2022},
url={https://openreview.net/forum?id=nZeVKeeFYf9}
}

@misc{gelato2025,
  title = {{Gelato} --- From Data Curation to Reinforcement Learning: Building a Strong Grounding Model for Computer-Use Agents},
  author = {Awadalla, Anas and Ghosh, Dhruba and Akkus, Aylin and Zhang, Yuhui and Nezhurina, Marianna and Jitsev, Jenia and Choi, Yejin and Schmidt, Ludwig},
  year = {2025},
  publisher = {GitHub},
  howpublished = {\url{https://github.com/mlfoundations/gelato}}
}

@article{guilibra,
  title={Gui-libra: Training native gui agents to reason and act with action-aware supervision and partially verifiable rl},
  author={Yang, Rui and Wu, Qianhui and Wang, Zhaoyang and Chen, Hanyang and Yang, Ke and Cheng, Hao and Yao, Huaxiu and Peng, Baolin and Zhang, Huan and Gao, Jianfeng and others},
  journal={arXiv preprint arXiv:2602.22190},
  year={2026}
}

@article{zwz2026,
  title={Zooming without Zooming: Region-to-Image Distillation for Fine-Grained Multimodal Perception},
  author={Wei, Lai and He, Liangbo and Lan, Jun and Dong, Lingzhong and Cai, Yutong and Li, Siyuan and Zhu, Huijia and Wang, Weiqiang and Kong, Linghe and Wang, Yue and others},
  journal={arXiv preprint arXiv:2602.11858},
  year={2026}
}

@article{loshchilov2019adamw,
  title={Decoupled weight decay regularization},
  author={Loshchilov, Ilya and Hutter, Frank},
  journal={arXiv preprint arXiv:1711.05101},
  year={2017}
}

@misc{easyocr,
  title = {{EasyOCR}},
  author = {{JaidedAI}},
  howpublished = {GitHub repository},
  url = {https://github.com/JaidedAI/EasyOCR},
  note = {Software. Accessed September 21, 2026},
  year = {2020}
}

@misc{gemini35flash,
  title = {{Gemini 3.5 Flash} Model Card},
  author = {{Google DeepMind}},
  year = {2026},
  month = may,
  url = {https://deepmind.google/models/model-cards/gemini-3-5-flash/},
  note = {Published May 19, 2026. Accessed September 21, 2026}
}

@inproceedings{regionfocus,
  title={Visual test-time scaling for gui agent grounding},
  author={Luo, Tiange and Logeswaran, Lajanugen and Johnson, Justin and Lee, Honglak},
  booktitle={2025 IEEE/CVF International Conference on Computer Vision (ICCV)},
  pages={19989--19998},
  year={2025}
}
\bibliographystyle{iclr2027_conference}

\appendix
\makeatletter
\setlength{\@fptop}{0pt}
\setlength{\@fpsep}{16pt plus 2pt minus 2pt}
\makeatother
\setcounter{table}{0}
\renewcommand{\thetable}{A\arabic{table}}
\renewcommand{\theHtable}{A\arabic{table}}

\section{Diagnostic Analysis Details}
\label{sec:app_diagnostic_details}

\subsection{Predictions and Element Labels}
\label{sec:app_prediction_corpus}
\begin{figure}[!htb]
    \centering
    \includegraphics[width=\textwidth]{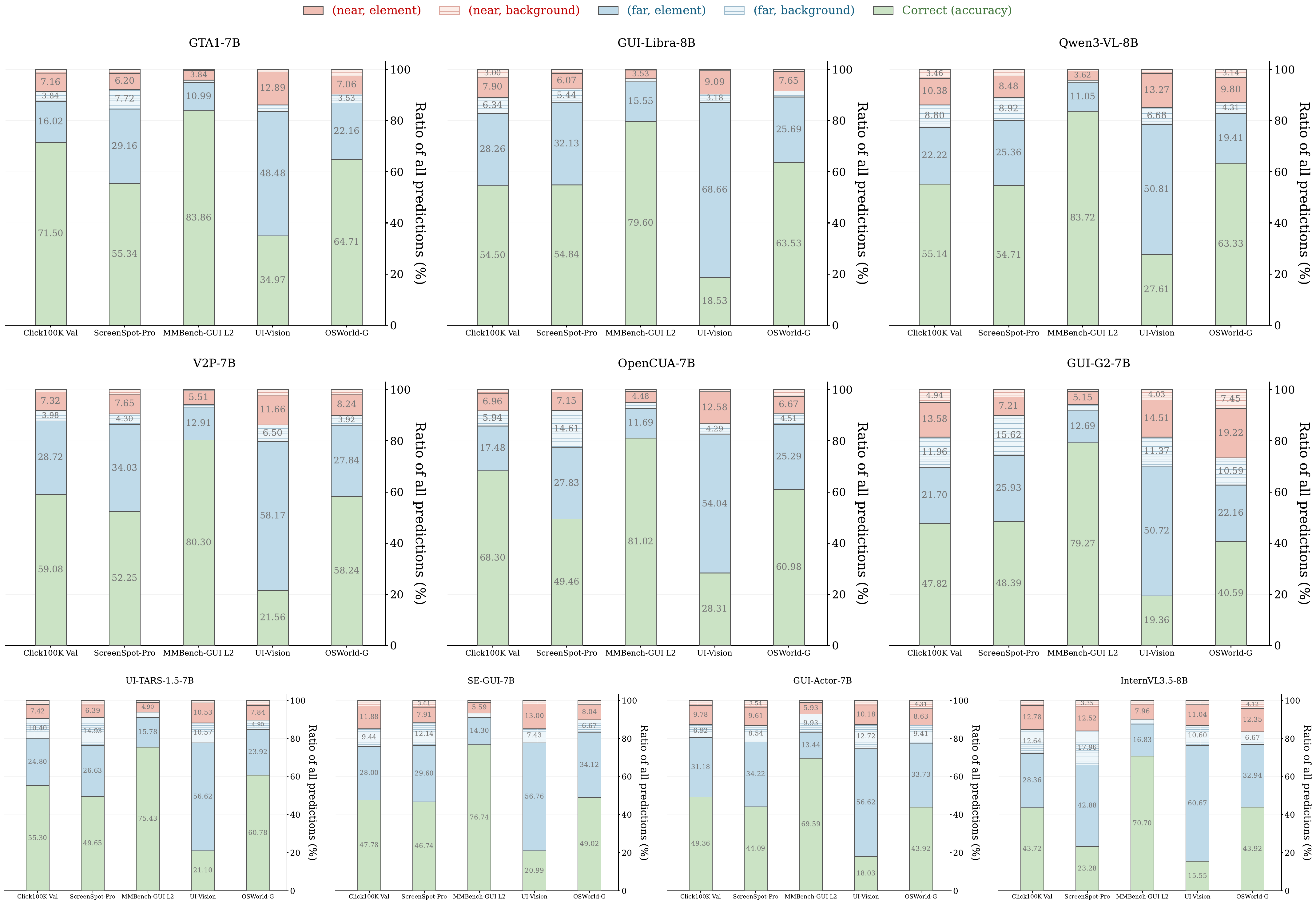}
    \caption{Prediction distributions for ten baseline models across five datasets. Each bar reports accuracy and four error rates over all predictions, with a $3w\times3h$ neighborhood. }
    \label{fig:app_per_model_distributions}
\end{figure}

The benchmark subsets described in Section~\ref{sec:error_decomposition_setup} contain 1,581 ScreenSpot-Pro examples, 3,594 MMBench-GUI L2 examples, 5,479 UI-Vision examples, and 510 OSWorld-G examples.
Fig.~\ref{fig:app_per_model_distributions} shows prediction distributions for all ten evaluated models.

\noindent\textbf{Parser settings.}
OmniParser-v2 icon detection uses confidence threshold 0.05 and intersection-over-union threshold 0.7.
The detector preserves aspect ratio, using the original longer side rounded up to a multiple of 32.
EasyOCR~\citep{easyocr} uses English recognition with text-detection threshold 0.8.
We retain text and icon boxes with area at least 16 pixels and at most 35\% of the screenshot area.
For shared-element analysis, a click in overlapping regions is assigned to the smallest containing region, with original region order breaking ties.
Element coverage is computed as the area of the union of retained regions.

\noindent\textbf{Manual inspection of parsing results.}
Two annotators inspect the detections on 100 randomly selected screenshots.
A screenshot meets the quality criterion when incorrectly detected elements account for at most 5\% of all GUI elements in that screenshot.
Missed elements are not counted as detection errors in this inspection, so the criterion assesses detection correctness rather than coverage.
Overall, 87\% of the inspected screenshots meet this criterion.

\noindent\textbf{Manual verification of non-target clicks.} 
To verify the accuracy of automated error categorization, two annotators manually reviewed 100 cases automatically identified as non-target clicks based on model predictions and parser results. Human inspection confirmed that 92\% (92/100) of these cases were true non-target clicks.

\noindent\textbf{Model-assisted review of distractor errors.}
\label{sec:app_distractor_review}
We randomly sample 500 screenshots from examples classified as element errors in the diagnostic analysis, where a grounding model clicks outside the ground-truth target region and on another parsed GUI element.
The review aims to distinguish actual selections of the wrong element from disagreements caused by incorrect ground-truth annotations or cases where both the annotated target and the model-selected element are valid choices, for example because multiple actions are acceptable or the instruction is ambiguous.
For each sampled example, we provide Gemini 3.5 Flash~\citep{gemini35flash} with the instruction, a clean screenshot, a marked full screenshot, and a marked zoomed view, using green for the annotated target and red for the model-selected element.
The prompt below asks Gemini to judge which regions match the instruction: only the annotated target (A), only the model-selected element (B), both (C), or neither (D).
Option A indicates an actual element-selection error, option B suggests an incorrect ground-truth annotation, and option C indicates that the model-selected element is also acceptable and should not be treated as a wrong-element selection.
Option D shows that neither candidate aligns with the instructions, leaving the target undefined.
Gemini selects option A in 94\% of cases reviewed, indicating that most instances are due to element-selection errors rather than annotation mistakes or valid alternatives.
The following prompt is provided to Gemini:

\begin{quote}
\small
\begin{minipage}{\linewidth}
You are checking a GUI grounding example.\\
The two candidate targets are the green bounding box and the red bounding box.

\medskip
Instruction: [instruction]

\medskip
Which candidate bounding box corresponds to the instruction?

\medskip
A: only the green bounding box corresponds to the instruction.\\
B: only the red bounding box corresponds to the instruction.\\
C: both bounding boxes correspond to the instruction, and either click would work.\\
D: neither bounding box corresponds to the instruction.\\
Return JSON only as \texttt{\{"answer":"A"\}}.
\end{minipage}
\end{quote}

\noindent\textbf{Counting conventions.}
In Fig.~\ref{fig:diagnostic_more}(b), the denominator at threshold $k$ counts examples with at least $k$ valid incorrect predictions.
The numerator counts examples with at least $k$ models selecting the same parsed element, counting each example once.
In Fig.~\ref{fig:diagnostic_more}(a), the observed element-hit proportion and random-click expectation use the same set of valid incorrect predictions for each model.


\subsection{Neighborhood Density and Scale Sensitivity}
\label{sec:app_neighborhood_density}
We test whether the higher error density near the target persists across models, datasets, and neighborhood scales.
We vary the neighborhood size $rw\times rh$ with $r\in\{2,3,4,5\}$ for all ten models and five datasets.

\begin{figure}[htbp]
\centering
\includegraphics[width=\textwidth]{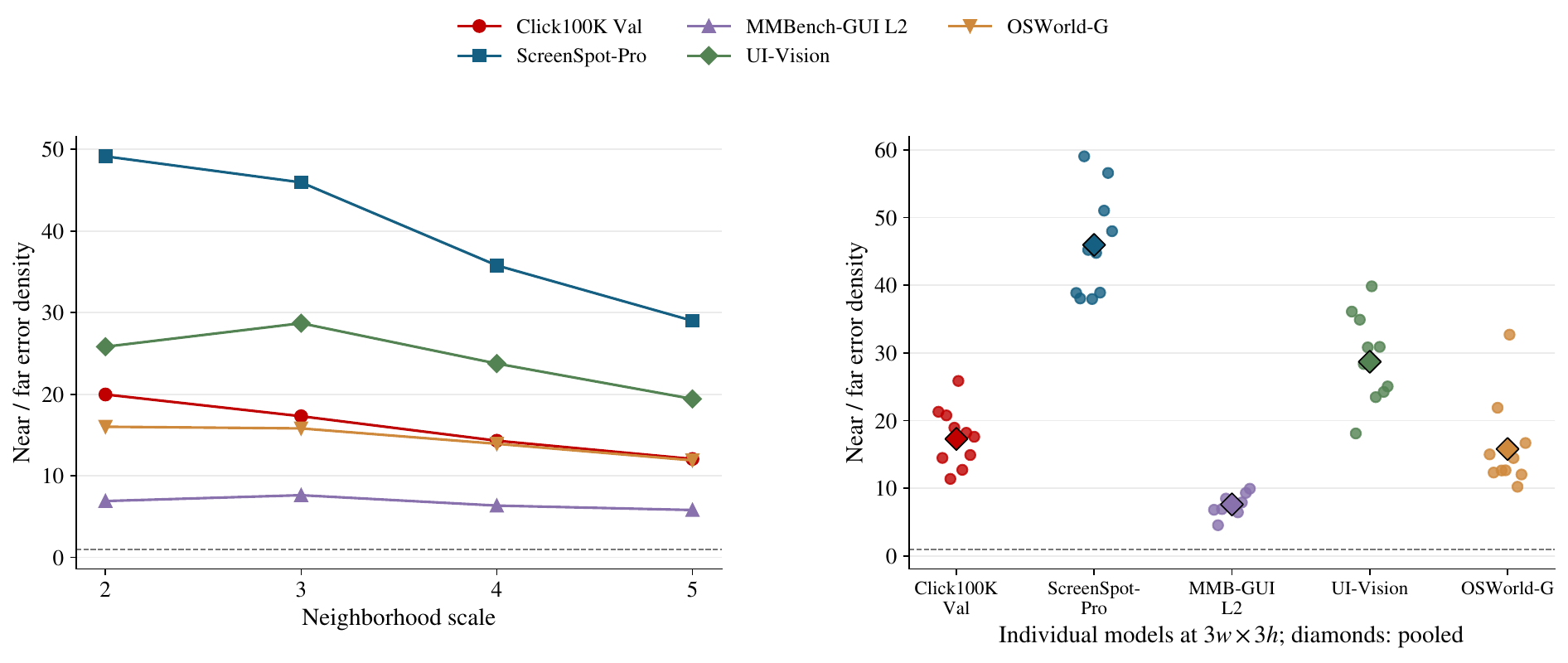}
\caption{Neighborhood-scale sensitivity across five datasets (left) and individual-model density ratios at $r=3$ (right). Small points represent ten models; diamonds represent pooled errors. The dashed line marks equal near and far density.}
\label{fig:app_density_sensitivity}
\end{figure}

For each example, we clip the target-centered neighborhood to the screenshot.
For polygon targets, $w$ and $h$ are the dimensions of the enclosing box.
Let $a_r$ be the neighborhood area as a fraction of screenshot area, averaged over all examples in a dataset and including the target area.
Let $e_r$ be the fraction of valid incorrect clicks inside the neighborhood.
We compute $e_r$ for each model separately and for predictions pooled across models.
The near/far error density ratio is
\begingroup
\fontsize{8pt}{9pt}\selectfont
\begin{equation}
R_r=\frac{e_r/a_r}{(1-e_r)/(1-a_r)}.
\end{equation}
\endgroup
A ratio above one indicates higher error density inside the neighborhood.
Fig.~\ref{fig:app_density_sensitivity} summarizes the results.
The ratio exceeds one in all 200 model--dataset--scale combinations, with a minimum of 2.70.
Thus, the higher near-target error density persists across the evaluated models, datasets, and neighborhood sizes.

\begin{figure}[htbp]
\centering
\includegraphics[width=\textwidth]{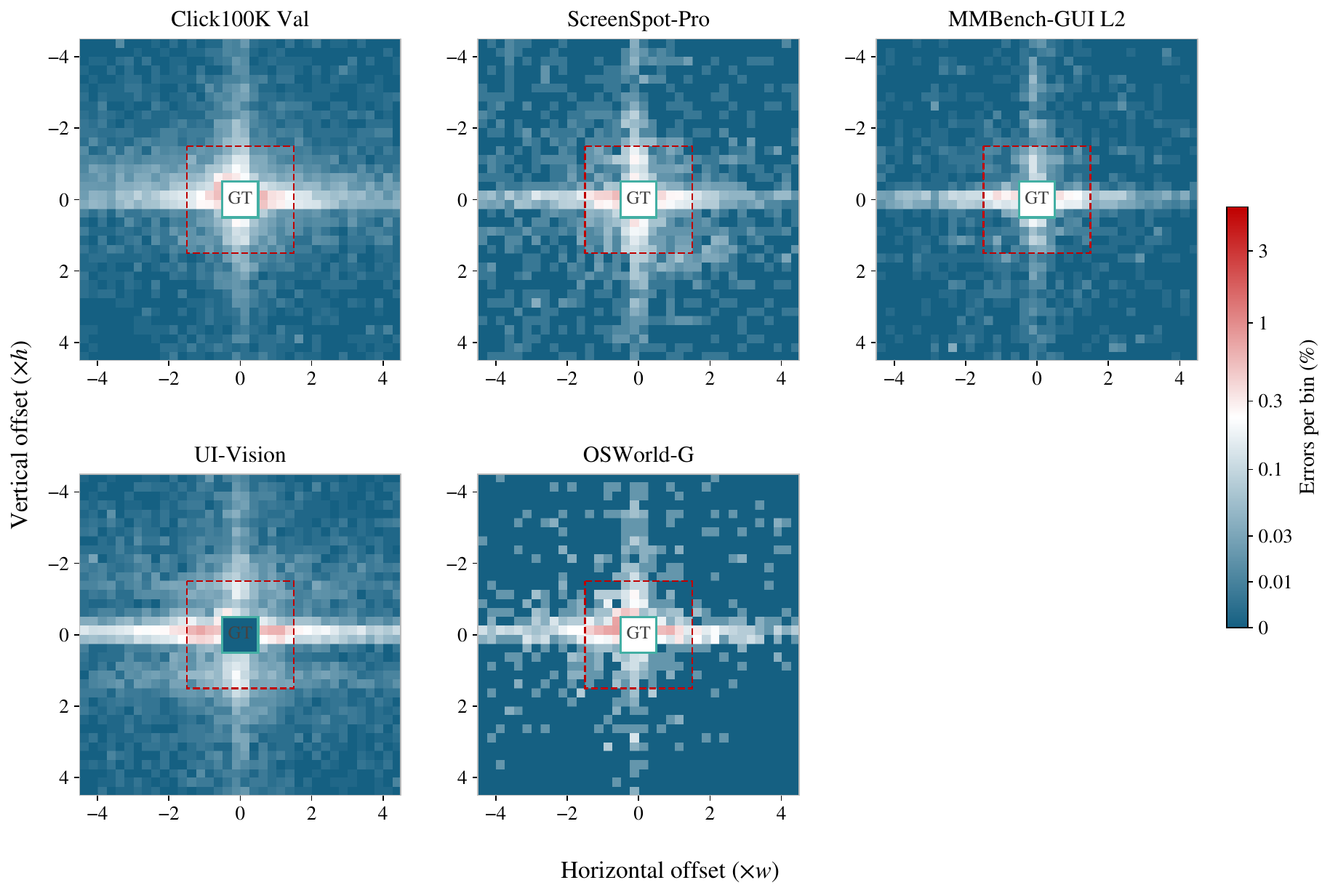}
\caption{Pooled spatial distributions of incorrect clicks across five datasets, extending the main-text heatmap. Colors give the percentage of all predictions per bin on a shared logarithmic scale; red denotes more errors.}
\label{fig:app_heatmaps_pooled}
\end{figure}

\begin{samepage}
\noindent\textbf{Spatial heatmaps.}
Fig.~\ref{fig:app_heatmaps_pooled} extends the main-text heatmap to all five datasets, pooling predictions across the ten models.
Offsets from the target center are normalized by target width and height, with bins of size $0.25\times0.25$.
Colors show incorrect clicks per bin as a percentage of all predictions, using a shared logarithmic scale.
The dashed box marks the $3w\times3h$ neighborhood.
\end{samepage}

\section{\modelname{} Data and Training Details}
\label{sec:app_semantic_data_details}

\noindent\textbf{Training data collection.}
\label{sec:app_candidate_verification}
We collect training examples from Click100K, using the original screenshots, instructions, and annotated target boxes.
For each example, GTA1-7B and GUI-Libra-8B predict a click location from the screenshot and instruction.
We match incorrect clicks to GUI element regions detected by OmniParser-v2.
An example becomes a candidate when both models select the same non-target element, and examples with multiple candidate distractors are discarded.
We pair the selected element box with the annotated target box. 

Agreement on a non-target element may reflect an incorrect ground-truth annotation.
We therefore use Gemini 3.5 Flash~\citep{gemini35flash} to inspect the screenshot and instruction and assess whether the ground-truth annotation is correct.
We retain only pairs for which this review confirms that the annotated target is correct and the shared model prediction is an incorrect selection.
This verification step yields 4,292 paired training examples.
To maintain coverage of the different Click100K subsets, we additionally sample ordinary grounding examples across the 86 source subsets using an equal per-subset sampling budget.
We match the total number of ordinary examples to the 4,292 mined target--distractor pairs, giving 8,584 training examples in total.
Our goal is to evaluate learning from observed distractors, not optimize data selection.

The following prompt specifies the pair-verification task, with the annotated target marked in green and the shared model-selected element marked in red:

\begin{quote}
\small
You are reviewing a candidate GUI grounding training pair.
Given the screenshot and instruction, determine whether the green-box element is a valid target and the red-box element is an incorrect choice.
The green box is the dataset annotation; the red box is the element selected by two grounding models.
Neither the annotation nor model agreement guarantees correctness.

\medskip
Instruction: [instruction]

\medskip
Judge each candidate using the visible interface and the instruction, including the requested action and any constraints on location or context.
Do not reject the red-box element merely because it differs from the annotation.
If both elements would independently satisfy the instruction, including when multiple actions are acceptable or the wording permits both choices, classify both as valid.
If the screenshot does not provide enough evidence to decide, report uncertainty rather than guessing.

\medskip
\begin{minipage}{\linewidth}
Choose exactly one option:\\
A: The green-box element satisfies the instruction, and the red-box element does not.\\
B: The red-box element satisfies the instruction, and the green-box element does not.\\
C: Both elements satisfy the instruction; either choice is acceptable.\\
D: Neither element satisfies the instruction.\\
E: There is insufficient evidence to determine which element satisfies the instruction.
\end{minipage}

\medskip
Return JSON only, with an \texttt{answer} field containing A, B, C, D, or E and a \texttt{reason} field briefly explaining the decision using evidence from the screenshot and instruction.
\end{quote}

Under this decision rule, only option A qualifies a candidate for retention as a target--distractor pair.

\noindent\textbf{Random training data.}
We retain Click100K training examples with complete instructions, image information, and target boxes, exclude the fixed 5,000-example validation set by row ID, and uniformly sample 8,584 examples without replacement from the remaining 96,313 candidates sorted by row ID, using random seed 20260808.

\noindent\textbf{Validation and evaluation.}
We use a held-out set of 5,000 Click100K examples, separate from the training data, for hyperparameter tuning and checkpoint selection.
This is the same Click100K validation set used for the error analysis in Section~\ref{sec:error_decomposition_setup}.
We also tune the hyperparameters of \method{} for the grounding models on this validation set.
The selected checkpoint is evaluated on ScreenSpot-Pro, MMBench-GUI L2, and OSWorld-G.
We omit UI-Vision from downstream evaluation because Click100K includes some UI-Vision data.


\noindent\textbf{Training configuration.} 
The following training recipe is for \modelname{}-8B, built on Qwen3-VL-8B.
We fine-tune the instruction-tuned backbone with LoRA rank 64, scaling factor 128, and dropout 0.05.
Adapters are applied to the attention and feed-forward projections, and the visual merger remains frozen.
We use AdamW~\citep{loshchilov2019adamw} with zero weight decay, gradient clipping at 1.0, and an effective batch size of 128.
Table~\ref{tab:app_training_setup} lists the stage-specific settings.
Refinement continues from the initial checkpoint with a new optimizer and learning-rate schedule.
 
\begin{table}[htbp]  
\centering  
\small
\setlength{\tabcolsep}{10pt}
\caption{Training settings for \modelname{}-8B. Both stages use cosine learning-rate decay.}
\label{tab:app_training_setup}
\begin{tabular}{lcc}
\toprule
Setting & Initial training & Refinement \\
\midrule
Learning rate & $3.45\times10^{-5}$ & $1\times10^{-6}$ \\
Schedule length (updates) & 320 & 30 \\
Warmup updates & 10 & 10 \\
Selected checkpoint (update) & 110 & 10 \\
Selection loss weight $\lambda_{\mathrm{sel}}$ & 0.1 & 0.05 \\
\bottomrule
\end{tabular}
\end{table}

The attention readout combines decoder layers 18--23 (zero-based) and is trained jointly with the adapters at learning rate $10^{-4}$.
The pairwise loss uses margin $m=0.3$ and weight $\lambda_{\mathrm{pair}}=0.5$.
Visual tokens are assigned to regions by image-patch overlap.
Tokens shared by the target and distractor masks are removed from both masks, and pairs with an empty mask contribute only coordinate loss.

\noindent\textbf{Direct inference.}
Coordinate-only training and \modelname{} use the same prompt requesting a point $[x,y]$ on a 0--1000 scale.
Images retain their aspect ratio, with a pixel budget between 3,136 and 8,847,360.
Decoding is greedy with a maximum of 32 generated tokens.
The pretrained Qwen3-VL-8B baseline uses its released computer-use prompt and pixel coordinates.

\subsection{Sensitivity to the Pairwise Margin}
\label{sec:app_margin_sensitivity}
\begin{figure}[htbp]
\centering
\includegraphics[width=\textwidth]{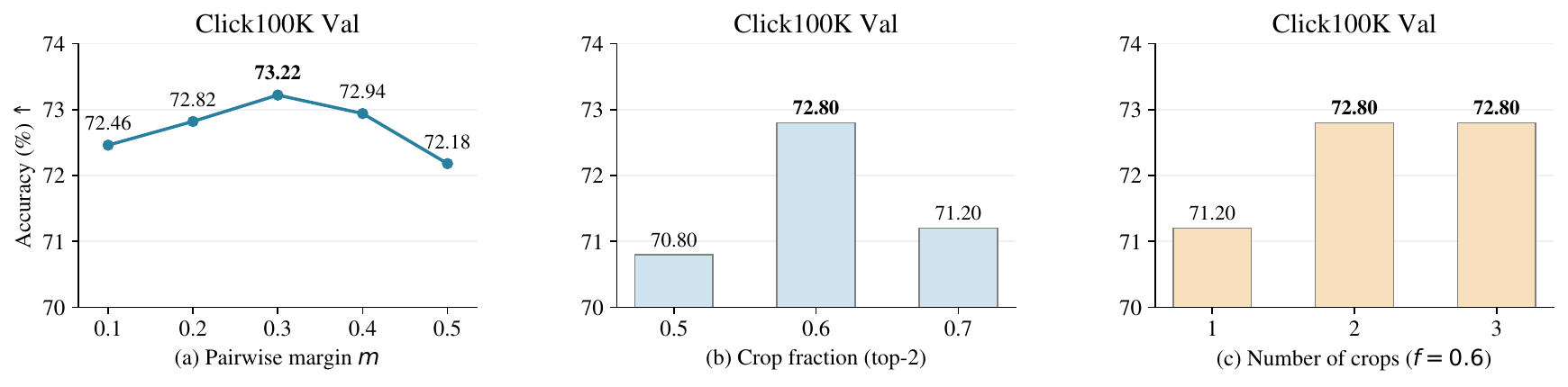}
\caption{Sensitivity to training and inference parameters. (a) Pairwise margin on Click100K Val. (b) \method{} crop fraction with two score-guided crops and (c) number of score-guided crops at crop fraction 0.6, evaluated on Click100K Val. Labels show accuracy (\%); bold marks the highest value in each panel.}
\label{fig:app_margin_sensitivity}
\end{figure}

We evaluate the pairwise margin $m\in\{0.1,0.2,0.3,0.4,0.5\}$ on Click100K Val.
Fig.~\ref{fig:app_margin_sensitivity}(a) shows accuracies between 72.18\% and 73.22\%, with the highest observed accuracy at $m=0.3$.

\subsection{Training Ablations on Additional Backbones}
\label{sec:app_additional_backbones}
Table~\ref{tab:app_additional_backbones} extends the training ablations to InternVL3.5-8B and Qwen3-VL-30B-A3B-Instruct.
For each backbone, we compare coordinate-only training on random data, coordinate-only training on our data, and training with the full objective.
Using our data improves accuracy on all three benchmarks for both backbones, and the full objective yields further improvements.

\begin{table*}[htbp]
\centering
\fontsize{8pt}{9pt}\selectfont
\setlength{\tabcolsep}{4.3pt}
\caption{Training ablations on additional backbones. All results are accuracy (\%). Darker green indicates better performance; bold marks the best results within each backbone.}
\label{tab:app_additional_backbones}
\begin{tabular}{cc l r|r|r|r}
\toprule
\multicolumn{2}{c}{Data} & \multicolumn{1}{c}{Training objective} & ScreenSpot-Pro & OSWorld-G & MMB-GUI L2 & Avg. \\
\cmidrule(lr){1-2}\cmidrule(lr){3-3}\cmidrule(lr){4-4}\cmidrule(lr){5-5}\cmidrule(lr){6-6}\cmidrule(lr){7-7}
Random & Our & & Acc.$\uparrow$ & Acc.$\uparrow$ & Acc.$\uparrow$ & Acc.$\uparrow$ \\
\midrule
\multicolumn{7}{l}{\textbf{InternVL3.5-8B}} \\
$\checkmark$ & & Coordinate-only & \cellcolor[HTML]{F2F8F4}25.62 & \cellcolor[HTML]{F2F8F4}46.67 & \cellcolor[HTML]{F2F8F4}71.45 & \cellcolor[HTML]{F2F8F4}47.91 \\
& $\checkmark$ & Coordinate-only & \cellcolor[HTML]{D2E7D9}25.81 & \cellcolor[HTML]{D2E7D9}47.65 & \cellcolor[HTML]{D2E7D9}72.65 & \cellcolor[HTML]{D2E7D9}48.70 \\
& $\checkmark$ & Full objective & \cellcolor[HTML]{98C9AA}\textbf{26.69} & \cellcolor[HTML]{98C9AA}\textbf{49.22} & \cellcolor[HTML]{98C9AA}\textbf{72.95} & \cellcolor[HTML]{98C9AA}\textbf{49.62} \\
\midrule
\multicolumn{7}{l}{\textbf{Qwen3-VL-30B-A3B-Instruct}} \\
$\checkmark$ & & Coordinate-only & \cellcolor[HTML]{F2F8F4}62.37 & \cellcolor[HTML]{F2F8F4}69.02 & \cellcolor[HTML]{F2F8F4}84.53 & \cellcolor[HTML]{F2F8F4}71.97 \\
& $\checkmark$ & Coordinate-only & \cellcolor[HTML]{D2E7D9}63.82 & \cellcolor[HTML]{D2E7D9}70.78 & \cellcolor[HTML]{D2E7D9}85.42 & \cellcolor[HTML]{D2E7D9}73.34 \\
& $\checkmark$ & Full objective & \cellcolor[HTML]{98C9AA}\textbf{64.77} & \cellcolor[HTML]{98C9AA}\textbf{72.55} & \cellcolor[HTML]{98C9AA}\textbf{85.61} & \cellcolor[HTML]{98C9AA}\textbf{74.31} \\
\bottomrule
\end{tabular}
\end{table*}

\section{\method{} Implementation and External Grounders}
\label{sec:app_inference_details}

\noindent\textbf{Crop configuration.}
For screenshot dimensions $W,H$ and crop fraction $f$, the crop width and height are
\begin{equation}
\begin{aligned}
w_f &= \min\bigl(W,\max(320,\operatorname{round}(fW))\bigr),\\
h_f &= \min\bigl(H,\max(320,\operatorname{round}(fH))\bigr).
\end{aligned}
\end{equation}
The initial-click crop uses $f=0.4$ and is centered on the initial prediction before boundary adjustment.
The four candidate grid crops use $f=0.6$, with normalized centers at $(0.3,0.3)$, $(0.7,0.3)$, $(0.3,0.7)$, and $(0.7,0.7)$.
Crops extending beyond the screenshot are shifted inside its boundaries while preserving the computed width and height.
Each selected crop is enlarged by $2\times$ using Lanczos resampling.
A visual token contributes to a crop score when its center lies inside the crop.
Fig.~\ref{fig:app_margin_sensitivity}(b,c) reports crop-parameter ablations on Click100K Val.
With two score-guided crops, crop fractions 0.5, 0.6, and 0.7 yield accuracies of 70.80\%, 72.80\%, and 71.20\%, respectively.
At crop fraction 0.6, selecting one, two, or three score-guided crops yields 71.20\%, 72.80\%, and 72.80\% accuracy, respectively.

\noindent\textbf{Candidate scoring.}
For candidate $p$ in view $v$, the coordinate log-likelihood $\ell_v(p)$ sums token log-probabilities over the native response format, excluding the end-of-sequence token.
Within each view, identical serialized coordinate responses are counted once when computing the mean $\mu_v$ and population standard deviation $\sigma_v$ over candidates whose locations lie in that view.
The standardized log-likelihood is
\begin{equation}
\bar\ell_v(p)=\frac{\ell_v(p)-\mu_v}{\max(\sigma_v,10^{-6})}.
\end{equation}
The generating view participates in this normalization, but cross-view support averages only the standardized scores from the other views containing the candidate, as in Eq.~\eqref{eq:lcr_cross}.
We then standardize cross-view support and spatial agreement separately across the remaining candidate set $\mathcal P'$ for the current example.
For either score $F\in\{S_{\mathrm{cross}},S_{\mathrm{agree}}\}$, this gives
\begin{equation}
\hat F(p)=\frac{F(p)-\mu_F}{\max(\sigma_F,10^{-6})},
\end{equation}
where $\mu_F$ and $\sigma_F$ are the mean and population standard deviation of $F$ over $\mathcal P'$.
Both stages floor the standard deviation at $10^{-6}$.
The final comparison maximizes $\hat S_{\mathrm{cross}}(p)+\hat S_{\mathrm{agree}}(p)$, as in Eq.~\eqref{eq:lcr_select}.
Ties are resolved in the order of the full-image prediction, the initial-click crop, and the score-guided crops ranked by region score.

\noindent\textbf{Inference baselines.}
On \modelname{}-8B, MVP uses attention from the coordinate-comma token at decoder layer 24 and makes up to two crop predictions.
It uses the same image budget as direct inference.
ZoomClick uses search depth 3, crop ratio 0.5, minimum crop size 768 pixels, a $2\times2$ initial grid, and a 50-pixel stopping threshold.
Crop centers are adjusted to keep crops inside the screenshot.

\subsection{Inference Comparisons and Ablations}
\label{sec:app_lcr_results}
Tables~\ref{tab:inference_comparison} and~\ref{tab:lcr_ablation} provide the complete three-benchmark results corresponding to Fig.~\ref{fig:lcr_analysis}(b) and (c), respectively.

\begin{table*}[t]
\centering
\fontsize{8pt}{9pt}\selectfont
\setlength{\tabcolsep}{4.3pt}
\caption{Comparison of inference methods across three backbones and benchmarks. Acc. denotes accuracy and E-err. denotes the element error rate, both in percentages. Within each backbone, \textcolor{blue}{\textbf{blue bold}} and \textbf{black bold} mark the best and second-best results, respectively.}
\label{tab:inference_comparison}
\begin{tabular}{llll|ll|ll}
\toprule
& & \multicolumn{2}{c}{ScreenSpot-Pro} & \multicolumn{2}{c}{OSWorld-G} & \multicolumn{2}{c}{MMB-GUI L2} \\
\cmidrule(lr){3-4}\cmidrule(lr){5-6}\cmidrule(lr){7-8}
\multicolumn{1}{l}{Backbone} & \multicolumn{1}{l}{Method} & \multicolumn{1}{l}{Acc.$\uparrow$} & \multicolumn{1}{l}{E-err.$\downarrow$} & \multicolumn{1}{l}{Acc.$\uparrow$} & \multicolumn{1}{l}{E-err.$\downarrow$} & \multicolumn{1}{l}{Acc.$\uparrow$} & \multicolumn{1}{l}{E-err.$\downarrow$} \\
\midrule
\multirow{3}{*}{UI-TARS-1.5-7B}
& MVP & \textbf{59.84} & \textbf{32.51} & \textbf{60.78} & \textbf{30.98} & \textbf{76.27} & \textbf{20.17} \\
& ZoomClick & 57.24 & 35.48 & 57.25 & 37.06 & 73.57 & 21.87 \\
& \method{} & \textcolor{blue}{\textbf{61.23}} & \textcolor{blue}{\textbf{30.55}} & \textcolor{blue}{\textbf{65.88}} & \textcolor{blue}{\textbf{30.59}} & \textcolor{blue}{\textbf{77.74}} & \textcolor{blue}{\textbf{19.59}} \\
\midrule
\multirow{3}{*}{GTA1-7B}
& MVP & \textbf{64.39} & \textbf{32.38} & \textbf{69.02} & \textbf{26.47} & \textbf{83.70} & \textbf{15.41} \\
& ZoomClick & 59.33 & 37.38 & 65.49 & 28.43 & 80.75 & 17.58 \\
& \method{} & \textcolor{blue}{\textbf{65.21}} & \textcolor{blue}{\textbf{30.30}} & \textcolor{blue}{\textbf{70.59}} & \textcolor{blue}{\textbf{25.10}} & \textcolor{blue}{\textbf{84.67}} & \textcolor{blue}{\textbf{14.50}} \\
\midrule
\multirow{3}{*}{\modelname{}-8B}
& MVP & \textbf{69.96} & \textbf{27.39} & \textbf{70.78} & \textbf{24.90} & \textbf{86.39} & \textbf{12.94} \\
& ZoomClick & 66.92 & 29.60 & 65.88 & 28.04 & 83.95 & 15.14 \\
& \method{} & \textcolor{blue}{\textbf{72.36}} & \textcolor{blue}{\textbf{25.87}} & \textcolor{blue}{\textbf{71.76}} & \textcolor{blue}{\textbf{23.73}} & \textcolor{blue}{\textbf{87.28}} & \textcolor{blue}{\textbf{12.10}} \\
\bottomrule
\end{tabular}
\end{table*}

\begin{table}[htbp]
\centering\fontsize{8}{9}\selectfont
\setlength{\tabcolsep}{4pt}
\caption{Component ablations of \method{}. Acc. and E-err. are percentages over all predictions. Avg. is the unweighted mean across the three benchmarks. All variants use the same \modelname{}-8B checkpoint. Bold marks the best value in each column.}
\label{tab:lcr_ablation}
\begin{tabular}{lrr|rr|rr|rr}
\toprule
& \multicolumn{2}{c}{ScreenSpot-Pro} & \multicolumn{2}{c}{OSWorld-G} & \multicolumn{2}{c}{MMB-GUI L2} & \multicolumn{2}{c}{Avg.} \\ \cmidrule(lr){2-3}\cmidrule(lr){4-5}\cmidrule(lr){6-7}\cmidrule(lr){8-9}
\multicolumn{1}{l}{Variant} & \multicolumn{1}{l}{Acc.$\uparrow$} & \multicolumn{1}{l}{E-err.$\downarrow$} & \multicolumn{1}{l}{Acc.$\uparrow$} & \multicolumn{1}{l}{E-err.$\downarrow$} & \multicolumn{1}{l}{Acc.$\uparrow$} & \multicolumn{1}{l}{E-err.$\downarrow$} & \multicolumn{1}{l}{Acc.$\uparrow$} & \multicolumn{1}{l}{E-err.$\downarrow$} \\
\midrule
\method{} & \textbf{72.36} & 25.87 & \textbf{71.76} & \textbf{23.73} & 87.28 & 12.10 & \textbf{77.13} & \textbf{20.57} \\
w/o score-guided crops & 66.29 & 30.42 & 69.80 & 25.10 & 86.28 & 13.05 & 74.12 & 22.86 \\
w/o learned crop selection & 72.17 & \textbf{25.68} & 70.59 & 24.90 & 87.26 & 12.27 & 76.67 & 20.95 \\
w/o $S_{\mathrm{cross}}$ & 70.97 & 26.31 & 71.57 & 23.92 & 87.15 & 12.19 & 76.56 & 20.81 \\
w/o $S_{\mathrm{agree}}$ & 72.23 & 26.12 & 71.18 & 24.51 & \textbf{87.48} & \textbf{12.08} & 76.96 & 20.90 \\
\bottomrule
\end{tabular}
\end{table}

\subsection{External Point Grounders}
\label{sec:app_external_lcr}

\noindent\textbf{Attention readouts.}
Table~\ref{tab:app_lcr_model_settings} summarizes the attention readout used by \method{} for each model.
All three use the last prompt token as the query.
For \modelname{}-8B, learned weights combine pre-softmax query--key scores across all 32 heads in layers 18--23.
For GTA1-7B and UI-TARS-1.5-7B, we obtain crop scores from frozen model attention.
GTA1 takes the maximum attention probability across all heads in layers 12--25.
These probabilities are normalized over all keys, restricted to image tokens, and converted to log space before region pooling.
UI-TARS uses pre-softmax query--key scores from head 25 in layer 24.
Layer and head indices are zero-based.
The readout for each model is selected by accuracy on the held-out Click100K validation set and fixed across all three benchmarks.
Crop construction and candidate scoring follow the procedure above.

\begin{table}[htbp]
\centering
\fontsize{8pt}{9pt}\selectfont
\setlength{\tabcolsep}{5pt}
\caption{Model-specific attention readouts for \method{}. QK denotes pre-softmax query--key scores. Layer and head indices are zero-based. External readouts use temperature 1. Each configuration is shared across all three evaluation benchmarks.}
\label{tab:app_lcr_model_settings}
\begin{tabular}{lllll}
\toprule
Model & Layers & Heads & Attention scores & Aggregation \\
\midrule
\modelname{}-8B & 18--23 & All 32 & QK & Learned weighted sum \\
GTA1-7B & 12--25 & All & Probabilities & Maximum \\
UI-TARS-1.5-7B & 24 & 25 & QK & Single head \\
\bottomrule
\end{tabular}
\medskip

\begin{tabular}{lc}
\toprule
Shared parameter & Value \\
\midrule
Initial-click crop fraction & 0.4 \\
Grid crop fraction & 0.6 \\
Number of score-guided crops & 2 \\
Minimum crop dimension & 320 pixels \\
Crop enlargement factor & $2\times$ \\
Cross-view support coefficient & 1 \\
Spatial agreement coefficient & 1 \\
\bottomrule
\end{tabular}
\end{table}

\noindent\textbf{Model interfaces and baselines.}
For UI-TARS, all methods use the released grounding prompt, a 12,845,056-pixel image budget, and greedy decoding, sharing the same initial prediction.
GTA1 uses method-specific point-model interfaces, so its initial predictions can differ between methods.
Candidate likelihoods use coordinates mapped to each resized view and the native response format of each model.
MVP uses coordinate-comma attention at layer 20 and up to four crop predictions for these external models.
Its ScreenSpot-Pro procedure is also used for MMBench-GUI L2, while OSWorld-G uses the corresponding benchmark-specific procedure.
The UI-TARS implementation adapts MVP attention extraction to the shared model interface.
ZoomClick uses the search settings specified above with a point-model adapter.
All methods use the same accuracy and element-error evaluation, with failed predictions included in the denominator.

\section{Detailed Benchmark Results}
\label{sec:app_benchmark_subsets}
Table~\ref{tab:app_ssp_subsets} reports accuracy and element error rates by domain on ScreenSpot-Pro, and Table~\ref{tab:app_mmb_subsets} reports subset accuracies on MMBench-GUI L2. Both use the same locally evaluated models and predictions as Table~\ref{tab:main_results}.

\begin{table*}[htbp]
\centering
\fontsize{7pt}{8pt}\selectfont
\setlength{\tabcolsep}{2pt}
\caption{Performance by domain on ScreenSpot-Pro (\%). Acc. is point accuracy; E-err. is the percentage of all predictions in each domain that fall outside the target region but inside a parsed GUI element. Each domain pools Text and Icon targets. Overall pools all examples and matches Table~\ref{tab:main_results}; all results use the same locally evaluated models and predictions. \srcverified{} retains the Table~\ref{tab:main_results} marker on overall accuracy. \textcolor{blue}{\textbf{Blue bold}} and \textbf{black bold} mark the best and second-best values in each column, respectively; ties receive the same marking.}
\label{tab:app_ssp_subsets}
\resizebox{\textwidth}{!}{%
\begin{tabular}{lrr|rr|rr|rr|rr|rr|rr}
\toprule
\multicolumn{1}{l}{Method} & \multicolumn{2}{c}{CAD} & \multicolumn{2}{c}{Dev} & \multicolumn{2}{c}{Creative} & \multicolumn{2}{c}{Scientific} & \multicolumn{2}{c}{Office} & \multicolumn{2}{c}{OS} & \multicolumn{2}{c}{Overall} \\
\cmidrule(lr){2-3}\cmidrule(lr){4-5}\cmidrule(lr){6-7}\cmidrule(lr){8-9}\cmidrule(lr){10-11}\cmidrule(lr){12-13}\cmidrule(lr){14-15}
\multicolumn{1}{l}{} & \multicolumn{1}{c}{Acc.$\uparrow$} & \multicolumn{1}{c}{E-err.$\downarrow$} & \multicolumn{1}{c}{Acc.$\uparrow$} & \multicolumn{1}{c}{E-err.$\downarrow$} & \multicolumn{1}{c}{Acc.$\uparrow$} & \multicolumn{1}{c}{E-err.$\downarrow$} & \multicolumn{1}{c}{Acc.$\uparrow$} & \multicolumn{1}{c}{E-err.$\downarrow$} & \multicolumn{1}{c}{Acc.$\uparrow$} & \multicolumn{1}{c}{E-err.$\downarrow$} & \multicolumn{1}{c}{Acc.$\uparrow$} & \multicolumn{1}{c}{E-err.$\downarrow$} & \multicolumn{1}{c}{Acc.$\uparrow$} & \multicolumn{1}{c}{E-err.$\downarrow$} \\
\midrule
GUI-Actor-7B & 31.03 & 52.11 & 39.13 & 46.49 & 43.70 & 43.99 & 48.03 & 46.46 & 60.87 & 28.26 & 44.90 & 43.37 & 44.09\srcverified{} & 43.83 \\
SE-GUI-7B & 41.76 & 41.38 & 43.81 & 42.81 & 37.83 & 39.88 & 54.72 & 34.65 & 70.43 & 22.17 & 35.20 & 41.84 & 46.74\srcverified{} & 37.51 \\
GUI-G2-7B & 48.28 & 37.55 & 46.49 & 38.13 & 40.47 & \textbf{36.36} & 53.54 & 31.10 & 68.70 & 20.00 & 34.69 & 32.14 & 48.39\srcverified{} & 33.14 \\
OpenCUA-7B & 32.95 & 39.85 & 44.82 & 41.14 & 45.16 & 42.23 & 58.66 & \textbf{29.53} & 71.74 & 22.17 & 47.96 & \textcolor{blue}{\textbf{28.57}} & 49.46\srcverified{} & 34.98 \\
UI-TARS-1.5-7B & 45.98 & 32.57 & 46.15 & 40.47 & 44.28 & 36.95 & 55.12 & 32.68 & 69.57 & 20.00 & 38.78 & 31.12 & 49.65\srcverified{} & 33.02 \\
V2P-7B & 49.81 & 45.21 & 46.49 & 44.48 & 43.11 & 51.61 & 57.48 & 37.40 & 73.48 & 23.48 & 48.47 & 42.35 & 52.25\srcverified{} & 41.68 \\
GUI-Libra-8B & 49.81 & 44.44 & 51.17 & 43.14 & 51.91 & 40.47 & 55.51 & 39.76 & 72.17 & 19.57 & 51.02 & 38.27 & 54.84\srcverified{} & 38.20 \\
GTA1-7B & 50.19 & 44.83 & 51.84 & 37.79 & 49.85 & 38.12 & 55.51 & 39.37 & 78.26 & 18.26 & 50.00 & \textbf{29.08} & 55.34\srcverified{} & 35.36 \\
\midrule
Qwen3-VL-8B (baseline) & 49.43 & 39.08 & 52.51 & 36.12 & 50.44 & 38.12 & 55.51 & 33.46 & 72.61 & 16.96 & 50.51 & 36.22 & 54.71 & 33.84 \\
\modelname{}-8B (Ours) & \textbf{65.13} & \textbf{29.89} & \textbf{63.88} & \textbf{32.44} & \textbf{56.01} & 38.71 & \textbf{66.14} & 30.71 & \textbf{85.65} & \textbf{12.61} & \textbf{64.80} & 33.16 & \textbf{66.03} & \textbf{30.30} \\
\modelname{}-8B + \method{} (Ours) & \textcolor{blue}{\textbf{71.26}} & \textcolor{blue}{\textbf{26.44}} & \textcolor{blue}{\textbf{69.57}} & \textcolor{blue}{\textbf{28.76}} & \textcolor{blue}{\textbf{65.40}} & \textcolor{blue}{\textbf{31.38}} & \textcolor{blue}{\textbf{72.44}} & \textcolor{blue}{\textbf{26.77}} & \textcolor{blue}{\textbf{89.57}} & \textcolor{blue}{\textbf{10.00}} & \textcolor{blue}{\textbf{69.90}} & \textcolor{blue}{\textbf{28.57}} & \textcolor{blue}{\textbf{72.36}} & \textcolor{blue}{\textbf{25.87}} \\
\bottomrule
\end{tabular}%
}
\end{table*} 

\begin{table*}[htbp]
\centering
\fontsize{7pt}{8pt}\selectfont
\setlength{\tabcolsep}{2pt}
\caption{Subset accuracy on MMBench-GUI L2 (\%). All results use the locally evaluated models and predictions from Table~\ref{tab:main_results}. Each platform is divided into Basic and Advanced (Adv.) instructions. Overall pools examples across all subsets; All matches the corresponding accuracy in Table~\ref{tab:main_results}. \textcolor{blue}{\textbf{Blue bold}} and \textbf{black bold} mark the best and second-best values in each column, respectively; ties receive the same marking.}
\label{tab:app_mmb_subsets}
\resizebox{\textwidth}{!}{%
\begin{tabular}{lrr|rr|rr|rr|rr|rr|rrr}
\toprule
Method & \multicolumn{2}{c}{Windows} & \multicolumn{2}{c}{macOS} & \multicolumn{2}{c}{Linux} & \multicolumn{2}{c}{iOS} & \multicolumn{2}{c}{Android} & \multicolumn{2}{c}{Web} & \multicolumn{3}{c}{Overall} \\
\cmidrule(lr){2-3}\cmidrule(lr){4-5}\cmidrule(lr){6-7}\cmidrule(lr){8-9}\cmidrule(lr){10-11}\cmidrule(lr){12-13}\cmidrule(lr){14-16}
 & Basic & Adv. & Basic & Adv. & Basic & Adv. & Basic & Adv. & Basic & Adv. & Basic & Adv. & Basic & Adv. & All \\
\midrule
GUI-Actor-7B & 61.99 & 55.51 & 59.71 & 59.83 & 49.74 & 41.84 & 90.45 & 82.73 & 70.79 & 79.72 & 89.68 & 72.08 & 71.80 & 67.40 & 69.59 \\
SE-GUI-7B & 79.34 & 56.25 & 75.07 & 58.67 & 65.45 & 44.39 & 95.22 & 83.64 & 94.66 & 85.35 & 90.97 & 71.10 & 84.89 & 68.68 & 76.74 \\
GUI-G2-7B & 80.44 & 56.99 & 81.45 & 66.76 & 68.06 & 48.47 & 95.86 & 82.42 & 95.79 & 84.51 & 92.90 & 76.95 & 87.24 & 71.39 & 79.27 \\
OpenCUA-7B & 90.77 & 62.13 & 84.06 & 65.03 & 81.68 & 58.16 & 95.22 & 80.00 & 95.22 & 79.72 & 94.19 & 76.30 & 90.77 & 71.39 & 81.02 \\
UI-TARS-1.5-7B & 76.75 & 56.62 & 73.62 & 60.98 & 70.16 & 51.53 & 90.45 & 80.00 & 92.42 & 80.00 & 85.48 & 72.40 & 82.48 & 68.46 & 75.43 \\
V2P-7B & 84.50 & 59.19 & 83.77 & 69.36 & 71.73 & 57.14 & 94.27 & 85.15 & 92.98 & 83.10 & 91.94 & 74.68 & 87.69 & 72.99 & 80.30 \\
GUI-Libra-8B & 86.72 & 63.60 & 86.38 & 70.23 & 64.92 & 48.98 & 91.08 & 80.30 & 94.38 & 83.10 & 91.61 & 73.38 & 87.47 & 71.83 & 79.60 \\
GTA1-7B & 89.67 & 68.01 & \textbf{88.99} & 71.39 & 79.58 & 60.71 & 95.86 & 82.42 & \textbf{96.91} & 84.79 & 94.84 & \textbf{80.52} & 91.89 & 75.93 & 83.86 \\
\midrule
Qwen3-VL-8B (baseline) & 87.82 & 67.65 & 86.96 & 68.79 & 75.39 & 61.22 & \textbf{96.82} & \textbf{86.36} & 96.63 & 86.20 & \textcolor{blue}{\textbf{96.45}} & 80.19 & 91.16 & 76.37 & 83.72 \\
\modelname{}-8B (Ours) & \textbf{91.51} & \textbf{69.85} & 86.96 & \textbf{78.61} & \textbf{82.20} & \textbf{66.84} & \textcolor{blue}{\textbf{97.13}} & \textcolor{blue}{\textbf{88.79}} & \textcolor{blue}{\textbf{97.47}} & \textcolor{blue}{\textbf{90.70}} & \textbf{95.81} & 77.60 & \textbf{92.56} & \textbf{80.08} & \textbf{86.28} \\
\modelname{}-8B + \method{} (Ours) & \textcolor{blue}{\textbf{93.73}} & \textcolor{blue}{\textbf{75.37}} & \textcolor{blue}{\textbf{90.72}} & \textcolor{blue}{\textbf{80.92}} & \textcolor{blue}{\textbf{82.72}} & \textcolor{blue}{\textbf{71.94}} & 96.50 & 86.06 & 95.51 & \textbf{87.89} & \textbf{95.81} & \textcolor{blue}{\textbf{81.17}} & \textcolor{blue}{\textbf{93.17}} & \textcolor{blue}{\textbf{81.46}} & \textcolor{blue}{\textbf{87.28}} \\
\bottomrule
\end{tabular}%
}
\end{table*}

\end{document}